\documentclass[sn-mathphys,Numbered]{sn-jnl}% Math and Physical Sciences Reference Style
\usepackage{graphicx}%
\usepackage{multirow}%
\usepackage{amsmath,amssymb,amsfonts}%
\usepackage{amsthm}%
\usepackage{mathrsfs}%
\usepackage[title]{appendix}%
\usepackage[table]{xcolor}%
\usepackage{textcomp}%
\usepackage{manyfoot}%
\usepackage{booktabs}%
\usepackage{algorithm}%
\usepackage{algorithmicx}%
\usepackage{algpseudocode}%
\usepackage{listings}%
\begin{document}

\title[Evaluating Explanatory Capabilities of Machine Learning Models...]{Evaluating Explanatory Capabilities of Machine Learning Models in Medical Diagnostics: A Human-in-the-Loop Approach}


\author[1]{\fnm{José} \sur{Bobes-Bascarán}}\email{jose.bobes@udc.es}

\author*[1]{\fnm{Eduardo} \sur{Mosqueira-Rey}}\email{eduardo@udc.es}

\author[1]{\fnm{Ángel} \sur{Fernández-Leal}}\email{angel.fleal@udc.es}

\author[1]{\fnm{Elena} \sur{Hernández-Pereira}}\email{elena.hernandez@udc.es}

\author[1]{\fnm{David} \sur{Alonso-Ríos}}\email{david.alonso@udc.es}

\author[1]{\fnm{Vicente} \sur{Moret-Bonillo}}\email{vicente.moret@udc.es}

\author[1]{\fnm{Israel} \sur{Figueirido-Arnoso}}\email{israel.figueirido.arnoso@udc.es}

\author[2]{\fnm{Yolanda} \sur{Vidal-\'Insua}}\email{yvidalinsua@gmail.com}

\affil*[1]{\orgdiv{Department of Computer Science and Information Technologies}, \orgname{University of Coruña (CITIC)}, \orgaddress{\street{Rúa Maestranza, 9}, \city{La Coruña}, \postcode{15001}, \state{Galicia}, \country{Spain}}}

\affil*[2]{\orgdiv{Servicio de Oncología Médica}, \orgname{Complejo Hospitalario (CHUS)}, \orgaddress{\street{R\'ua da Choupana, s/n}, \city{Santiago de Compostela}, \postcode{15706}, \country{Spain}}}


\abstract{
This paper presents a comprehensive study on the evaluation of explanatory capabilities of machine learning models, with a focus on Decision Trees, Random Forest and XGBoost models using a pancreatic cancer dataset. We use Human-in-the-Loop related techniques and medical guidelines as a source of domain knowledge to establish the importance of the different features that are relevant to establish a pancreatic cancer treatment. These features are not only used as a dimensionality reduction approach for the machine learning models, but also as way to evaluate the explainability capabilities of the different models using agnostic and non-agnostic explainability techniques.
To facilitate interpretation of explanatory results, we propose the use of similarity measures such as the Weighted Jaccard Similarity coefficient. The goal is to not only select the best performing model but also the one that can best explain its conclusions and aligns with human domain knowledge.
}




\maketitle

%%%%%%%%%%%%%%%%%%%%%%%%%%%%%%%%%%%%%%%%%%%%%%%%%%%%%%%%%%%%%%%%%%%%%%%%%%%%%%%%
\section{Introduction}
\label{sec:introduction}
%%%%%%%%%%%%%%%%%%%%%%%%%%%%%%%%%%%%%%%%%%%%%%%%%%%%%%%%%%%%%%%%%%%%%%%%%%%%%%%%

% Explainable AI
Explainable AI (XAI) \cite{linardatos2021explainable} is a research field focused on making Artificial Intelligence (AI) systems in general, and Machine Learning (ML) systems in particular, more understandable to humans. 
% Advantages of XAI
Explainable AI offers several advantages, to name a few: it fosters confidence in the prediction of the model by making the decision-making process more transparent, promotes responsible AI development, aids in debugging and identifying issues, and allows auditing of AI models and checking if they adhere to regulatory standards.

% An overview of the evolution of explainability
The inherent explainability of AI systems has not remained static but has changed considerably as a result of technological progress. In fact, explainability has become an increasingly difficult issue to tackle, as the internal functioning of AI systems has become less intelligible as they have become more complex \cite{angelov2021explainable}. 

% Initial models were transparent
Initially, symbolic AI models were explainable \textit{per se}, e.g., rule-based expert systems could easily show to their users which rules they had followed to make a given decision, even though the rules can incorporate measures of uncertainty and imprecision as, for example, in fuzzy systems. 
These type of AI models are considered \textit{transparent}, which means that the model itself is understandable \cite{barredo2020explainable}, being understandability the characteristic of a model to make a human understand its function without any need for explaining its internal structure or the algorithmic means by which the model processes data internally \cite{montavon2018methods}.

% Symbolic AI vs Machine Learning
However, rule-based systems presented too many limitations that hindered their development: they were rigid systems, with poor scalability and limited learning capabilities. To overcome the problems associated with symbolic AI, machine learning models were developed. These models can learn from data and improve their performance over time without being explicitly programmed. They are obviously very dependent on data to obtain their results, but, in general, their performance, scalability, and generalization features are superior to those present on symbolic models.

% trade-off interpretability and model performance
But these machine learning models often lack interpretability. Some of them can be considered transparent, such as Logistic Regression or Decision Trees, but these are usually the simplest models that offer lower performance in complex domains. More complex models such as Random Forest, Support Vector Machines, Neural Networks, and models based on Deep Learning (convolutional networks, recurrent networks, etc.) are opaque models. We can see that there seems to be a trade-off between model interpretability and model performance. More interpretable models tend to perform less well than less interpretable models.

% the goal is to obtain XAI with the same performance as DL
Therefore, the main question is not whether it is possible to arrive at explainable solutions---this has been true for a long time---but whether it is possible to obtain explainable solutions with an accuracy that is comparable to the highly effective but comparably opaque Deep Learning systems \cite{angelov2021explainable}.

% Agnostic models
In an attempt to include explainability into the opaque models, ``agnostic'' techniques have been developed, which can be applied seamlessly to any ML model, regardless of its inner processing or internal representations \cite{barredo2020explainable}. The problem with these agnostic or post-hoc explainability methods is their reliability. Some authors \cite{slack2021advances} pointed out that they are inconsistent, unstable, and provide very little insight into their correctness and reliability, in addition to being computationally inefficient.
It is therefore necessary to develop some way of validating the explanatory capabilities obtained by these models, especially in complex domains, such as medical environments, where it may not be easy to determine the correct answer to a given problem.

% Contribution
In this article, we face a problem related to pancreatic cancer, deciding the best treatment for a patient given his or her clinical characteristics. Since our dataset is tabular in nature and does not have too many cases, we decided to try to solve the problem using classical machine learning models, such as Decision Trees (DT) \cite{kotsiantis2013decision}, Random Forest (RF) \cite{breiman2001random}, or XGBoost models \cite{chen2016xgboost}. We have already seen that DT are transparent models, but RF and XGBoost are opaque. 

Since we are in a medical domain, we understand that the models must offer adequate explainability, but how can we validate the results? A first way would be to apply different methods of explainability on the same model and compare the results with each other. But we want to go further and check if the explainability of the models has medical significance, which would allow us to have more trust in them.

To verify the medical significance, we can rely on the opinions of medical experts and also use medical guidelines that act as a ``gold standard'' in the field. The former option involves some kind of \textit{Human-in-the-Loop} (HITL) approach \cite{mosqueira2023human}, which is often a complex process and require the active participation of human experts. The latter option relies on bibliography and manuals, which tend to be exhaustive and sometimes hard to navigate through. 

The contribution of this paper is, in addition to the comparison between models of explainability, the development of different ways of evaluation of explainability models based on the collaboration of human experts and the guidelines available in the domain and its application to a pancreatic cancer treatment problem.

The structure of the paper is as follows: section \ref{sec:state_art} describes the state of the art of explainable AI and briefly explains our current domain: pancreatic cancer. Section \ref{sec:data-and-methodology} describes the dataset used and the methodology that includes a description of the feature selection process, the different ML models used and the different XAI methods considered. Section \ref{sec:results} describes the results obtained by each ML model using different feature sets in terms of accuracy but also in terms of interpretability. Finally, we end up with some discussion (Section \ref{sec:discussion}) and conclusions (Section \ref{sec:conclusions}).

%%%%%%%%%%%%%%%%%%%%%%%%%%%%%%%%%%%%%%%%%%%%%%%%%%%%%%%%%%%%%%%%%%%%%%%%%%%%%%%%
\section{State of the art}
\label{sec:state_art}
%%%%%%%%%%%%%%%%%%%%%%%%%%%%%%%%%%%%%%%%%%%%%%%%%%%%%%%%%%%%%%%%%%%%%%%%%%%%%%%%

%-------------------------------------------------------------------------------
\subsection{Explainability}\label{sec:explainability}
%-------------------------------------------------------------------------------

% Concept of explainability and other synonyms
Explainable AI is often defined as a group of techniques that make it possible for human users to understand, appropriately trust, and effectively manage machine learning systems.

% Clarification
The terms ``explainability'' and ``interpretability'' have been often used in the literature as synonyms. However, researchers such as Gilpin et al. \cite{gilpin2018explaining} argue that explainability goes beyond mere interpretability. Explainable models are interpretable by nature, but the opposite is not necessarily true. To be explainable, models must be auditable and capable of defending their actions and offering relevant responses to inquiries. In this regard, authors such as Lipton \cite{lipton2018mythos} seek to refine the discourse on interpretability and question the oft-made assertions that linear models are interpretable and that deep neural networks are not. 

% Taxonomies of explainability methods
When analyzing explainability methods we have to take into account how to characterize these methods. A first division can be made regarding \textit{scope} and \textit{generality} \cite{carrillo2021individual}. 
% Scope: Local vs Global
In \textbf{scope} we can classify the methods as:
\begin{itemize}
    \item \textbf{Local}: These aim to understand why the model made a certain prediction for a given instance or group of nearby instances.
    \item \textbf{Global}: These aim to understand the behavior of the model as a whole. 
\end{itemize}

% Generality: Model-Agnostic vs Model-Specific
Regarding \textbf{generality}, we can classify methods as:
\begin{itemize}
    \item \textbf{Model-Agnostic}: These methods are independent of the specific ML model used and can be applied to any model, regardless of its architecture or underlying algorithms. They use techniques such as perturbing input features and observing the impact on model predictions. %Examples include LIME (Local Interpretable Model-agnostic Explanations) \cite{ribeiro2016why} and SHAP (SHapley Additive exPlanations) \cite{lundberg2017unified} that we will comment later on.   
    \item \textbf{Model-Specific}: These methods are tailored to a particular type of ML model and exploit the model’s internal structure and characteristics to provide explanations. The complexity of these methods depends on how transparent the models are.
\end{itemize}    

% Explainability leads to trust 
Building trust is a key pillar when creating AI systems intended to work in collaboration with humans. Without the understanding on how those systems make their decisions, humans will not trust them \cite{goodman2017european}, especially in domains such as defense, medicine, finance, or law, where trust is an essential aspect. As pointed out by Adadi and Berrada \cite{adadi2018peeking}, entrusting important decisions to a system that cannot explain itself presents obvious dangers.

%...............................................................................
\subsubsection{Explainability in healthcare}
%...............................................................................

% Explainability in medicine
In the specific domain of our research, that is, healthcare, Holzinger et al. \cite{holzinger2019causability} remarked that if medical professionals are complemented by sophisticated AI systems, they should have the means to understand the machine decision process. Also they investigated the requirements for building explainable AI systems for the medical domain in \cite{holzinger2017what}. First of all, we must emphasize that, given the criticality of the domain, humans must be able to understand and actively influence the decision processes. Regulatory aspects must also be considered (e.g., the GDPR of the European Union), as well as legal ones (e.g., the written text of medical reports is legally binding). Holzinger et al. \cite{holzinger2017what} argue that for this domain at least, the only way forward appears to be the integration of knowledge-based approaches (due to their interpretability) and neural approaches (due to their high efficiency). In particular, authors recommend hybrid distributional models combining sparse graph-based representations with dense vector representations, linking to lexical resources and knowledge bases.

% Specific examples of explainability in medicine.
When citing examples of explainability, we can refer to a review of the XAI literature of the last decade on healthcare by Loh et al. \cite{loh2022application}, as well as some recent specific cases where it has been applied on medical ML models, for example: a diagnostic classification model for brain tumor detection \cite{nair2023building}, a multi-label classification of electrocardiograms \cite{ganeshkumar2021explainable}, and a breast cancer survival model \cite{moncada2021explainable}. More examples for the medical imaging field using deep learning cancer detection models \cite{gulum2021review}, and for skin cancer recognition are described \cite{hauser2022explainable}.

% General review of explainability in medicine.
In fact, the vast majority of high-impact literature on XAI started to appear in 2018, since that was when interest in XAI really began to take off. In the reviewed literature, the most used technique by far was SHapley Additive exPlanations (SHAP), followed by Gradient-weighted Class Activation Mapping (GradCAM). SHAP is especially popular for traditional ML models, and Loh et al. \cite{loh2022application} conclude that ``SHAP has the potential to be used in healthcare by analyzing the contribution of biomarkers or clinical features (players) to a speciﬁc disease outcome (reward)''. The next most used technique was GradCAM, which, unlike SHAP, was overwhelmingly popular for Deep Learning models and suitable for visual explanations. The rest of the techniques found in the literature were: LIME, Layer-wise relevance propagation (LRP), Fuzzy classiﬁers, Explainable boosting machine (EBM), Case-based reasoning (CBR), rule-based systems, and other heatmap and saliency map generation methods \cite{loh2022application}. 

%...............................................................................
\subsubsection{Problems with explainability methods}
%...............................................................................

But all this work on explainability is not free of criticism. In the introductory section we already commented on how authors such as Slack et al. \cite{slack2021advances} pointed out that explainability methods are inconsistent, unstable, and provide very little insight into their correctness and reliability, in addition to being computationally inefficient. In particular, authors demonstrated that local explanation methods are computationally inefficient as they typically require a large number of black box model queries to construct local approximations. This can be prohibitively slow especially in the case of complex neural models.

Other authors such as Bhatt et al. \cite{bhatt2020explainable}, conduct a study of how explainability techniques are used by organizations that deploy ML models in their workflows. They are focused on the most popularly deployed local explainability techniques: feature importance, counterfactual explanations, adversarial training, and influential samples. These techniques explain individual predictions, making them typically the most relevant form of model transparency for end users. The authors found that while ML engineers are increasingly using explainability techniques as sanity checks during the development process, there are still significant limitations to current techniques that prevent their use to directly inform end users. These limitations include the need for domain experts to evaluate explanations, the risk of spurious correlations reflected in model explanations, the lack of causal intuition, and the latency in computing and showing explanations in real time.

In Dimanov et al. \cite{dimanov2020you}, the authors show a straightforward method for modifying a pre-trained model to manipulate the output of many popular feature importance explanation methods with little change in accuracy, thus demonstrating the danger of trusting such explanation methods. This work demonstrates that many popular explanation methods used in real-world settings are unable to reliably indicate whether a model is fair.

All these approaches work on the model to demonstrate their findings but there are also recent works that demonstrate the downsides of the post-hoc explanation techniques modifying the inputs \cite{ghorbani2019interpretation, slack2020fooling, dombrowski2019explanations}. In this sense, perturbation-based explanation methods such as LIME and SHAP are subject to additional problems: results vary between runs of the algorithms \cite{alvarez2018robustness, lee2019developing, fen2019why, zafar2019dlime}, and hyperparameters used to select the perturbations can greatly influence the resulting explanation \cite{fen2019why}.  

This leads us to wonder how reliable these explainability techniques are and to what extent we can rely on them in complex environments such as medical settings. As commented by Bhatt et al. \cite{bhatt2020explainable} in these scenarios, it is necessary to rely on the collaboration of human experts and to check domain standards (medical guidelines) to verify that both the models and the explainability mechanisms are doing the right thing.

%-------------------------------------------------------------------------------
\subsection{Application domain: pancreatic cancer} 
%-------------------------------------------------------------------------------

Pancreatic cancer is the seventh leading cause of global cancer deaths in industrialized countries. It produced approximately half a million cases and caused almost the same number of deaths (4.5\% of all deaths caused by cancer) in 2018 \cite{bray2018global}.

% Brief introduction to pancreatic cancer. 
Pancreatic ductal adenocarcinoma is the most common form of pancreatic cancer, making up more than 80\% of cases. The disease begins in the cells of the pancreas ducts, which transport juices containing digestive enzymes into the small intestine.

Risk factors include having a family history of the disease, history of chronic inflammation of the pancreas (pancreatitis), Lynch syndrome, diabetes, being overweight or obese, and smoking \cite{tomczak2015cancer}.

Nowadays, surgical resection is the only treatment that offers a potential cure for pancreatic cancer, and the addition of chemotherapy in the adjuvant setting has been shown to improve survival rates \cite{mcguigan2018pancreatic}.

The application of AI in the healthcare domain, and particularly to pancreatic cancer is wide spreading in recent years. It is highly demanded and could be applied in different scenarios, ranging from early cancer diagnosis \cite{hunter2022role} \cite{kenner2021artificial}, classification of lesion \cite{dmitriev2021visualAnalytics}, or survival prediction \cite{bakasa2021survivalPrediction} \cite{walczak2017evaluation}. 

Although there are many successful examples of AI applied to pancreatic cancer \cite{hayashi2021recent}, several challenges remain unsolved, such as the lack of sufficient volume of data on the experiments carried out by different research teams to statistically validate their results \cite{bradley2019personalized}.

%...............................................................................
\subsubsection{Cancer staging}\label{sec:cancer_staging}
%...............................................................................

Cancer staging is the process of determining the extent of cancer in a patient's body and ascertaining its precise location. It allows to establish the severity of the cancer based on the main tumor, and also to diagnose if there are other organs affected by the cancer.

The correct diagnosis of cancer and its staging are key factors in understanding the patient's situation and establishing a treatment plan. It also serves as a basis for communication between physicians and patients. The most extended staging system worldwide is called TNM Staging, and it has been proposed by the American Joint Committee on Cancer (AJCC) \cite{amin2017eighth} and the Union for International Cancer Control (UICC) to be the standard staging system \cite{cong2018tumor}.

The meaning of T, N and M for pancreatic cancer is explained below:
\begin{itemize}
    \item \textbf{The extent of the tumor (\textbf{T})}: How large is the tumor and has it grown outside the pancreas into nearby blood vessels?
    \item \textbf{The spread to nearby lymph nodes (\textbf{N})}: Has the cancer spread to nearby lymph nodes? If so, how many of the lymph nodes have cancer?
    \item \textbf{The spread (metastasized) to distant sites (\textbf{M})}: Has the cancer spread to distant lymph nodes or distant organs such as the liver, peritoneum (the lining of the abdominal cavity), lungs, or bones?
\end{itemize}

Furthermore, based on the clinical stage of the main tumor, pancreatic cancer is classified into four types: 
\begin{itemize}
    \item \textbf{Stage 0}, cancer is present but it has not spread.
    \item \textbf{Stage I (no spread or resectable)}, cancer is limited to the pancreas and has grown 2 cm (stage IA) or its size is greater than 2 cm but less than 4 cm (stage IB).
    \item \textbf{Stage II (local spread or borderline resectable)}, the cancer is limited to the pancreas and its size is greater than 4 cm, or there is spread locally to the nearby lymph nodes.
    \item \textbf{Stage III (wider spread or unresectable)}, cancer may have expanded to nearby blood vessels or nerves but has not metastasized to distant sites.
    \item \textbf{Stage IV (metastatic)}, cancer has spread to distant organs.
\end{itemize}

This classification creates a common understanding in between the medical experts, and provide us with several features to build ML models based on them. 

%...............................................................................
\subsubsection{Pancreatic cancer medical guidelines}\label{sec:pancreatic_cancer_guidelines}
%...............................................................................

In this work, we have used the ``Pancreatic Adenocarcinoma - NCCN Clinical Practice Guidelines in Oncology'' \cite{nccn2022pancreatic}, which is one of the most widely used guidelines by oncologists for the diagnosis and treatment of pancreatic cancer. These guidelines are organized in processes (PANCs) that can be performed at different stages of diagnosis and treatment, for example when there is a clinical suspicion of pancreatic cancer (PANC-1), when the cancer is resectable (PANC-2), when the cancer is borderline resectable (PANC-3), or unresectable (PANC-4) and so on. 

\begin{figure}[htbp!]
\centering
\includegraphics[width=1.0\textwidth]{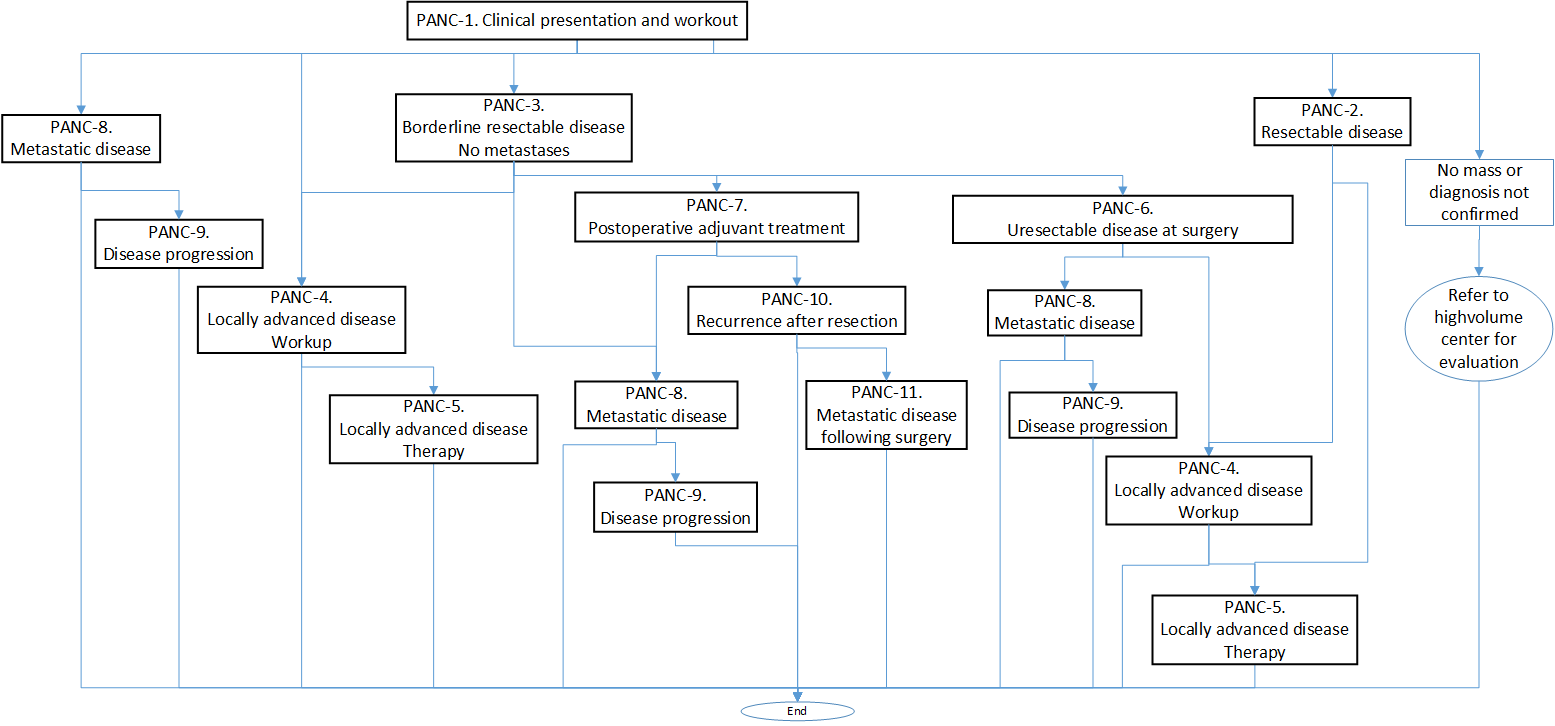}
\caption{Calling scheme between the different processes (PANCs).}
\label{fig:guidelines_PANCs_diagram}
\end{figure}

Each \textit{PANC} is organized in a sort of algorithm represented by a flowchart with the different tasks to be performed. For example, in Figure \ref{fig:guidelines_PANCs_diagram} we can see the calling scheme between different PANCs, and in Algorithm \ref{alg:cap} we can see a summary of the contents of PANC-1.

\begin{algorithm}
\caption{Clinical presentation and workup (PANC-1) \cite{nccn2022pancreatic}.}\label{alg:cap}
\begin{algorithmic}
\State - Clinical suspicion of pancreatic cancer or evidence of dilated pancreatic and/or bile duct (Indicators: age, gender)
\State - Pancreatic protocol CT (abdomen)
\State - Multidisciplinary consultation
\If{No metastatic disease}
	\State Medical tests:
	\State * Chest and pelvic CT
	\State * Consider endoscopic ultrasonography (EUS)
	\State * Consider MRI as clinically indicated for indeterminate liver lesions
	\State * Consider PET/CT in high-risk patients
	\State * Consider endoscopic retrograde cholangiopancreatography (ERCP) with stent placement
	\State * Liver function test and baseline CA 19-9 after adequate biliary drainage
	\State * Genetic testing for inherited mutations if diagnosis confirmed 
	\State Possible outcomes:
	\State * Refer to high-volume center for evaluation
	\State * Resectable Disease, Treatment (Indicators: primary diagnosis, site of resection or biopsy)	
	\State * Borderline Resectable Disease, No Metastases	(Indicators: primary diagnosis, site of resection or biopsy)
	\State * Locally Advanced Disease (Indicators: primary diagnosis, site of resection or biopsy)
	\State * Metastatic Disease, First-Line Therapy, and Maintenance Therapy
\ElsIf{Metastatic disease}
    \State Metastatic Disease
    \State Biopsy confirmation, from a metastatic site preferred
    \State Medical test:
    \State * Genetic testing form inherited mutations
    \State * Molecular profiling of tumor tissue is recommended
    \State * Complete staging with chest and pelvic CT
    \State Outcomes: (Indicators: primary diagnosis, site of resection or biopsy)
    \State * Metastatic Disease: First-Line and Maintenance Therapy 
\EndIf	  
\end{algorithmic}
\end{algorithm}

% Typical treatments
Following this guideline we can observe in the procedure corresponding to the clinical presentation and workup the four main diagnostic groups that are used to establish the treatment to be followed and that correspond with the stages presented in the Cancer Staging subsection (\ref{sec:cancer_staging}):

\begin{itemize}
    \item \textbf{Resectable disease}. In this case the guidelines suggest proceeding with surgery (without neoadjuvant therapy), or EUS-guided biopsy if neoadjuvant therapy is considered and if not previously done, and consider stenting if clinically indicated.
    \item \textbf{Borderline resectable disease}. Here a EUS-guided biopsy is preferred (if not previously done), and a staging laparoscopy and Baseline CA is considered. 
    \item \textbf{Locally advanced disease}. In this case a biopsy should be performed if not previously done, which may result in: a) cancer not confirmed, b) adenocarcinoma confirmed, and c) other cancer confirmed.
    \item \textbf{Metastatic disease}. Here the guidelines suggest: a) if jaundice present: placement of self-expanding metal stent, b) Genetic testing for inherited mutations, if not previously done, and c) Molecular profiling of tumor tissue, if not previously done. 
\end{itemize}

In this context, it should be borne in mind that many of the diagnostic tests require a multidisciplinary consultation including an appropriate set of imaging studies, which implies that the diagnostic and treatment process is often subject to the heuristic knowledge of oncologists.

Algorithm \ref{alg:cap} also shows a first approximation to the main indicators in the steps of the diagnosis and treatment process. In addition, after the interdisciplinary consultation, in the case of non-metastatic disease, the size and extent of the tumor (pathologic T), the number of nearby lymph nodes that are cancerous (pathologic N), the presence of metastases (pathologic M) and, from these, the pathological status are considered in the treatment of unresectable disease at surgery (for cases of resectable disease and borderline resectable disease), and in the treatment of locally advanced disease.

%%%%%%%%%%%%%%%%%%%%%%%%%%%%%%%%%%%%%%%%%%%%%%%%%%%%%%%%%%%%%%%%%%%%%%%%%%%%%%%%
\section{Data and methodology}
\label{sec:data-and-methodology}
%%%%%%%%%%%%%%%%%%%%%%%%%%%%%%%%%%%%%%%%%%%%%%%%%%%%%%%%%%%%%%%%%%%%%%%%%%%%%%%%

%-------------------------------------------------------------------------------
\subsection{Dataset}
%-------------------------------------------------------------------------------

The dataset used in this work was obtained from The Cancer Genome Atlas Program \cite{CancerGenomeAtlasResearchNetwork:2013}, a joint effort between the USA National Cancer Institute (NCI) and the National Human Genome Research Institute. The Cancer Genome Atlas (TCGA) is a landmark cancer genomics program that sequenced and molecularly characterized over 11,000 cases of primary cancer samples.

One of the projects included in the program is called TCGA-PAAD, focused on pancreatic cancer, which provides 185 diagnosed cases with the necessary details to carry out its complete analysis. For example, it includes information such as: the “stage event” which describes the pathological state of the tumor, “clinical data” which describes the characteristics of the tumor, and the occurrence of “new tumor events” which describes the patient follow-up. We have only considered 181 cases since the remaining four were incomplete. From the 181 cases considered, 117 were applied a chemotherapy treatment. Each of the cases has a total of 158 features, ranging from family history data to treatment and follow-up information.

%-------------------------------------------------------------------------------
\subsection{Feature selection}\label{sec:feature_selection}
%-------------------------------------------------------------------------------

For our experiment we did a meticulous feature selection process in which only those features that added relevant information to the use case of whether to apply a chemotherapy treatment were selected \cite{mcguigan2018pancreatic}. 

As a first step, we analyze the dataset with all available information. Even if the data provided by the PAAD repository allows to upload many variables for each patient case, the projects that had collaborated with the institution did not provide all the information possible.

The second step was a pruning process performed over those features without information or with very few informed cases. We removed unnecessary data as the PAAD database includes many variables that do not intervene in the treatment decision process. Moreover, some information provided on the dataset was redundant, as for example the size of the tumor which can be inferred from the TNM diagnosis (as commented in section \ref{sec:cancer_staging}), so we eliminated the unnecessary data.

Finally, we asked a panel of oncologists, who participated as human experts, selecting only those features they considered relevant to treatment selection in pancreatic cancer patients. Our goal in including human experts in the feature selection process is to improve the explanatory power of the resulting models. As described in \cite{mosqueira2024addressing}, they could also be involved with the aim of obtaining models with a higher accuracy. 

After the expert selection and pruning process, we ended up with 27 features and the target variable which is the therapy type. From this group of features we created three different sets of features --- \textit{recommended}, \textit{maximum}, and \textit{minimum} --- that would be used in the different ML models developed.

%...............................................................................
\subsubsection{Recommended set}
%...............................................................................

The inclusion of expert knowledge in the training of ML models can help to obtain better results and improve their interpretability. In this work, we first consider the collaboration of experts during the feature engineering process. We asked a panel of  medical doctors to rate the relevance regarding the prescription of chemotherapy treatment, of the 27 features selected during the feature selection process. Each feature should receive a value between 1 and 3, being 1 highly relevant, 2 relevant, and 3 barely relevant.

From the results provided by the medical experts, we extracted a set of recommended features that includes those features that received a value of 1 or 2. Among the features selected by the experts, three of them were dependent on the treatment prescribed, so we also decided to remove them from the set as the objective is treatment prescription. They were namely the primary therapy outcome success, new tumor events, and days to new tumor event after the initial treatment.
The details of the 14 features selected, the expert value received, and a brief description extracted from the GDC Data dictionary \footnote{\url{https://docs.gdc.cancer.gov/Data_Dictionary/viewer/}} are shown in Table \ref{tab:recommended-set-of-features}.

\begin{table}[htbp!]
\caption{\textit{Recommended} set of features.}\label{tab:recommended-set-of-features}%
\begin{tabular}{@{}llp{7.5cm}@{}}
\toprule
Feature                 & Experts    & Description \\
\midrule
Age                     & 2         & The patient’s age (in years). \\
Adenocarcinoma invasion & 1         & Confirm that the pancreas tumor sample being submitted to TCGA is an invasive adenocarcinoma.\\
Histological type       & 1         & Indicate the histologic subtype, if available, for the pancreas adenocarcinoma tumor sample being submitted to TCGA.\\
Initial diagnosis method & 2         & Initial pathologic diagnosis method \\
Lymph nodes positive HE & 1         & Number of lymph nodes positive by Hematoxylin and Eosin (HE) Stain. \\
Maximum tumor dimension & 1         & Provide the length of the largest dimension/diameter of the original tumor as stated on the pathology report.\\
Neoplasm cancer status  & 2         & The state or condition of an individual’s neoplasm at a particular point in time.\\
Neoplasm histologic grade&1         & A description of a tumor based on how abnormal the cancer cells and tissue look under a microscope and how quickly the cancer cells are likely to grow and spread. \\
Pathologic stage        & 1         & The extent of a cancer, especially whether the disease has spread from the original site to other parts of the body. \footnotemark[1] \\
Pathologic N            & 2         & Codes to represent the stage of cancer based on the nodes present. \footnotemark[1] \\
Pathologic M            & 1         & Code to represent the defined absence or presence of distant spread or metastases to locations via vascular channels or lymphatics beyond the regional lymph nodes.\footnotemark[1] \\
Residual tumor          & 1         & Text terms to describe the status of a tissue margin following surgical resection. \\
Surgery performed type  & 2         & Indicate the type of surgical procedure performed.\\
Year of initial diagnosis&1        & Year of initial diagnosis.\\
\botrule
\end{tabular}
\footnotetext[1]{Using criteria established by the American Joint Committee on Cancer (AJCC).}
\end{table}

%...............................................................................
\subsubsection{Maximum set}
%...............................................................................

From this point, we explore two different options. On the one hand, we enhance the feature set adding all the available information gathered (i.e., all 27 features). Even if some features were ``barely relevant'' for medical doctors, there are studies in which some evidence is assigned to them as risk factors for pancreatic cancer, as it is the case of gender, race, or ethnicity \cite{samaan2023pancreatic, nccnClinicalPractice2021}. We named the complete set of features as the \textit{maximum} set.

The idea behind this was to retain all the information available for the development of the ML models, so they can detect complex relationships among the features and ultimately produce a robust model. Therefore, we added 13 more features to our \textit{recommended} set (shown in Table \ref{tab:maximum-set-of-features}) ending up with the complete feature set.

\begin{table}[htbp!]
\caption{Features added to the \textit{maximum} set (jointly with those presented in Table \ref{tab:recommended-set-of-features}).}
\label{tab:maximum-set-of-features}%
\begin{tabular}{@{}llp{7.5cm}@{}}
\toprule
Feature                             & Expert    & Description \\
\midrule
Alcoholic exposure category         &           & Indicate the patient’s current level of exposure to alcohol. \\
Days to new tumor after treatment   & 1         & Days to new tumor after initial treatment. \\
Family history of cancer            & 3         & Indicate if a first degree relative (parents, siblings, or children) of the patient has a history of a cancer diagnosis. \\
Ethnicity                           &           & An individual’s self-described social and cultural grouping, specifically whether an individual describes themselves as Hispanic or Latino.\footnotemark[2] \\
Gender                              &           & Text designations that identify gender. Gender is described as the assemblage of properties that distinguish people on the basis of their societal roles.  \\
History of diabetes                 & 3         & Indicate if the patient has been previously diagnosed with diabetes. \\
New tumor events                    & 1         & Indicate whether a new tumor event occurs.  \\
Other DX                            &           & Numeric value to express the degree of abnormality of cancer cells, a measure of differentiation and aggressiveness. \\
Pathologic T                        & 3         & Code of pathological T (primary tumor) to define the size or contiguous extension of the primary tumor (T).\footnotemark[1]  \\
Primary therapy outcome success     & 1         & Indicates a complete remission or response to the prescribed treatment.\\
Race                                &           & An arbitrary classification of a taxonomic group that is a division of a species. It usually arises as a consequence of geographical isolation within a species and is characterized by shared heredity, physical attributes and behavior, and in the case of humans, by common history, nationality, or geographic distribution.\footnotemark[2] \\
Radiation therapy                   &           & Whether the treatment includes a radiation therapy.  \\
Tobacco smoking history             &           & Category describing current smoking status and smoking history as self-reported by a patient. \\
\botrule
\end{tabular}
\footnotetext[1]{Using criteria established by the American Joint Committee on Cancer (AJCC).}
\footnotetext[2]{Based on the categories defined by the U.S. Office of Management and Business and used by the U.S. Census Bureau.}
\end{table}

%...............................................................................
\subsubsection{Minimum set}
%...............................................................................

The other option chosen was to reduce the dimensionality of the dataset as the staging TNM variables implicitly include valuable information about the status of the cancer (see section \ref{sec:cancer_staging}). It is the case for example of the pathologic T that informs about the size of the tumor, or the pathologic M that indicates if a metastasis has occurred. Therefore, we considered the \textit{minimum} set of features as a dimensionality reduction with the goal of transforming high-dimensional data into a lower-dimensional representation without losing any important information. We ended up with five features (Age, T, N, M and Stage) that we can see in Table \ref{tab:minimum-set-of-features} along their range of values.

\begin{table}[htbp!]
\caption{\textit{Minimum} set with the range of values.}\label{tab:minimum-set-of-features}%
\begin{tabular}{@{}l p{1cm} p{1cm} p{1cm} p{3cm}@{}}
\toprule
Age & T & N & M & Stage  \\
\midrule
35-88 
& TX\newline T1\newline T2\newline T3\newline T4 
& N0\newline N1\newline N1b\newline NX 
& M0\newline M1\newline MX
& Stage 0\newline
Stage I, IA, IB\newline
Stage II, IIA, IIB\newline
Stage III\newline
Stage IV\\
\botrule
\end{tabular}
\end{table}

%...............................................................................
\subsubsection{Features and guidelines}
%...............................................................................

As a final step to evaluate feature importance we decided to follow the medical guidelines from the ``Pancreatic Adenocarcinoma - NCCN Clinical Practice Guidelines in Oncology'' \cite{nccn2022pancreatic} to see how they consider each feature in terms of importance. As we have seen in Figure \ref{fig:guidelines_PANCs_diagram} and Algorithm \ref{alg:cap}, these guidelines are quite complex and detail the sequential management decisions and interventions regarding pancreatic cancer, providing recommendations for some of the key issues in cancer prevention, screening, and care.

Therefore, feature importance regarding the chemotherapy treatment is not clearly specified in those guidelines so it is necessary to carry out a process of knowledge elicitation that allows us to extract such information. For this purpose, we analyzed the different procedural sheets into which the guideline is divided, taking into account that the same sheet can be considered in different diagnostic lines. This implies that successive tests may involve refinements or changes in determining the diagnosis and treatments to be used. 

\begin{figure}[htbp!]
\centering
\includegraphics[width=1.0\textwidth]{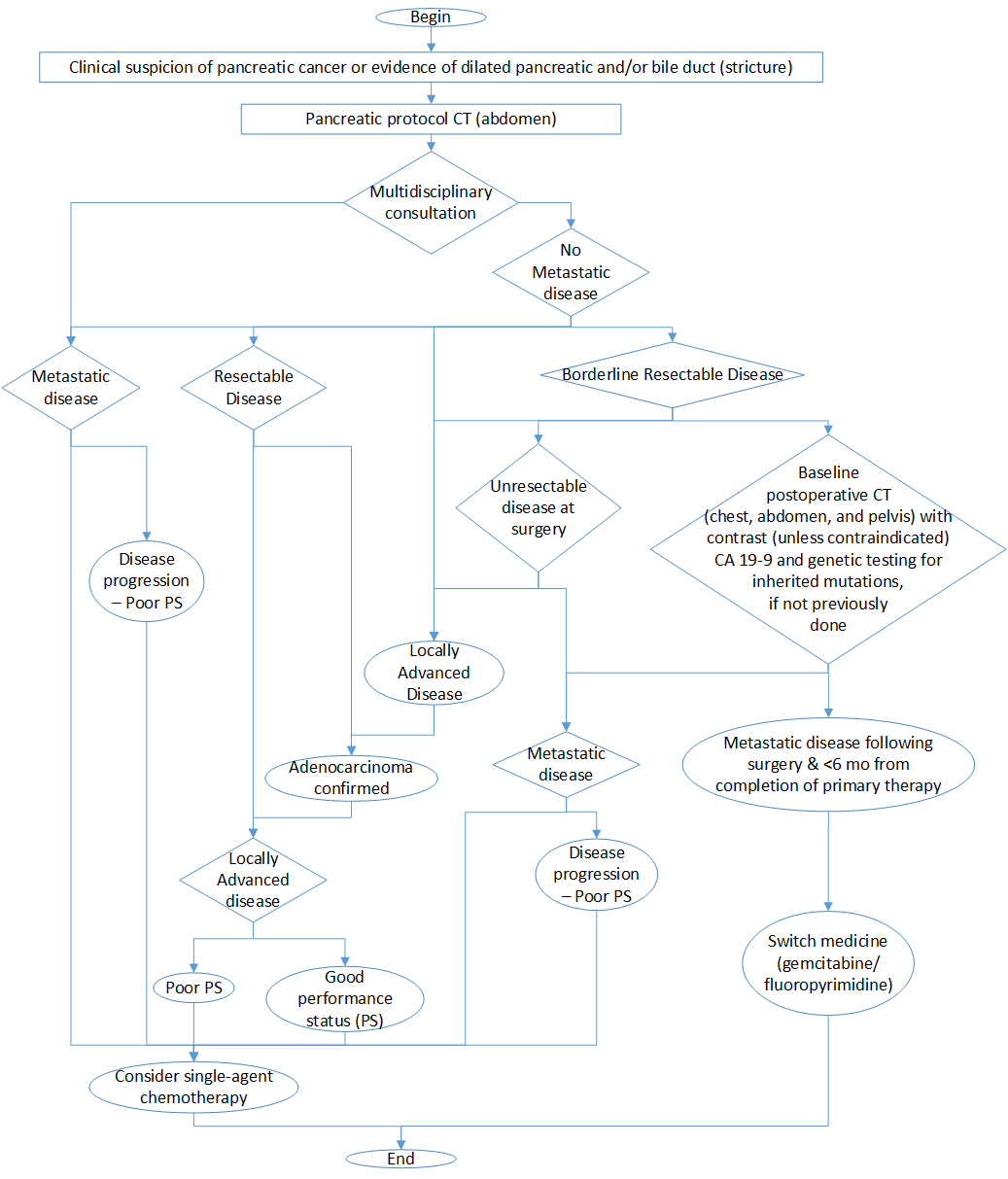}
\caption{Simplified graph of diagnostic decisions involving chemotherapy treatment.}
\label{fig:guidelines_chemotherapy_diagram}
\end{figure}

The final result of this analysis is presented in Figure \ref{fig:guidelines_chemotherapy_diagram} in which we show the decisions that we have to take leading to a chemotherapy treatment. It is important to remark that this figure is a simplification to highlight the fundamental decisions and considerations involved in chemotherapy treatment, and that the complete diagnostic process is more complex.

Each of the nodes represent an evaluation that we have roughly mapped onto a feature value, so that we can follow the diagnosis path. We could use the staging M as an evaluation of Metastatic disease, the size of the tumor T to determine whether it is resectable or not, and the N to evaluate if the cancer is locally advanced or not.

In Table \ref{tab:table_maximum_guidelines_experts} we can see the set of \textit{maximum} features ranked by the medical experts, as commented before, but also ranked using the information extracted from the medical guidelines.

\begin{table}[htbp!]
\centering
\footnotesize
\caption{Feature importance from the expert criteria and medical guidelines.}
\label{tab:table_maximum_guidelines_experts}
\begin{tabular}{l p{3.5cm} c c}
\toprule
Abbr. & Feature                     & Guidelines    & Experts \\
\midrule
Age     & Age                         & 3             & 2 \\
Stage   & Stage                       & 1             & 1 \\
T       & T                           & 2             & 3 \\
N       & N                           & 2             & 2 \\
M       & M                           & 2             & 1 \\
\midrule
Year    & Year of initial diagnosis   & 3             & 1 \\
Adeno.  & Adenocarcinoma invasion     & 2             & 1 \\
Type    & Histological type           & 2             & 1 \\
Status  & Neoplasm cancer status      & 1             & 2 \\
Grade   & Neoplasm histologic grade   & 3             & 1 \\
Dimensi.& Maximum tumor Dimension     & 1             & 1 \\
Residual& Residual tumor              & 3             & 1 \\
Diagnos.& Initial diagnosis method    & 2             & 2 \\
Surgery & Surgery performed type      & 2             & 2 \\
Lymph   & Lymph nodes positive by HE  & 1             & 1 \\
\midrule
Gender  & Gender                      &               &   \\
Race    & Race                        &               &   \\
Ethnic. & Ethnicity                   &               &   \\
Other   & Other DX                    &               &   \\
Diabetes& History of diabetes         & 3             & 3 \\
Family  & Family history of cancer    & 2             & 3 \\
Radiat. & Radiation therapy           & 2             & 2 \\
Therapy & Therapy outcome success     & 2             & 1 \\
N. tumor& New tumor events            &               & 1 \\
Days to & Days to new tumor           &               & 1 \\
Tobacco & Tobacco smoking history     &               &   \\
Alcohol & Alcoholic exposure category &               &   \\
\bottomrule
\end{tabular}
\end{table}

%-------------------------------------------------------------------------------
\subsection{Machine learning models}
%-------------------------------------------------------------------------------

With the aim of comparing different models in terms of accuracy and understandability, we chose three types of ML models: a \textbf{Decision Tree}, a \textbf{Random Forest} and an \textbf{XGBoost}. All of them are tree-based algorithms that are known to be easily understandable, but as they increase in complexity they require to be enhanced with ad-hoc exaplainability methods to be able to interpret their results. We do include a brief description of all of them highlighting the advantages and disadvantages in terms of accuracy and understandability.

As a general rule, when efficiency improvement is needed, aggregation techniques as bagging or bootstrapping are applied.
\begin{itemize}
    \item \textbf{Bagging}, also known as bootstrap aggregation, is the ensemble learning method that is commonly used to reduce variance within a noisy data set by combining the prediction of single models using various aggregation techniques.
    \item \textbf{Bootstrapping} is a technique used to iteratively improve a classifier's performance. Typically, multiple classifiers will be trained on different sets of the input data, and on prediction tasks the output of the different classifiers will be combined.
\end{itemize}
 
These methods are based on tree ensembles, which combine different single trees to obtain an aggregated prediction. Although the results are usually more efficient and mitigate overfitting, the interpretation of these models is far more complex. 

%...............................................................................
\subsubsection{Decision trees}
%...............................................................................

A Decision Tree (DT) is a non-parametric supervised learning algorithm, which is used for both classification and regression tasks. It has a hierarchical tree structure, which consists of a root node, branches, internal nodes and leaf nodes \cite{breiman2017classification}.

% Advantages of Decision Trees
DTs offer advantages over other AI models as they are easy to interpret. They do not require advance knowledge on AI to understand how they are built as they are a common structure used in many other fields. They allow a graphical representation of the algorithm, which makes them a perfect candidate in scenarios where understandability is key. Another significant advantage in these models is that they require a minimum set of data and the features are directly included in the model. The calculations required to get the a decision are also simple and easy to understand. They can handle both numerical and categorical data, and inherently perform feature selection.

% Disadvantages of Decision Trees
Nevertheless, they can produce overfitting, not generalizing competently when fed with new data. They are also sensitive to small changes in the input data, so small variations can generate very different trees. And finally, they are not good at discovering complex relationships between features.

To overcome these issues, simple models can be combined to get an aggregated model. However, and as usual, increasing the complexity of the model makes it harder to understand. Therefore, to be able to explain the decisions performed by the model, and infer which are the most relevant features taken into consideration, a complementary method is required.

%...............................................................................
\subsubsection{Random Forests}
%...............................................................................

Random Forests (RFs) are the result of combining a set of DTs forming ensembles. Each DT output is combined to reach a single result \cite{breiman2001random}. It operates by constructing a multitude of decision trees during the training phase and outputs the mode of the classes of the individual trees.

% Advantages of Random Forest
The main advantages of the RFs include high accuracy, robustness to overfitting, and the capacity to handle large datasets. In terms of understandability, they provide a way to measure the feature importance. Another great advantage is that RFs do not make assumptions about the distribution of the data and are able to capture complex relationships between features and the target variable. They scale in both number of features or observations, and their computation could be parallelized to get faster training times.

% Disadvantages of Random Forest
In simple scenarios, due to the complexity of the structure, it requires more computation than single trees, and its interpretability decrease. Even though the algorithm required to compute predictions can be parallelized, they are computationally intensive and consume more memory than individual decision trees.

%...............................................................................
\subsubsection{XGBoost}
%...............................................................................

XGBoost stands for “eXtreme Gradient Boosting” and it is built combining many simple models to create an ensemble with higher prediction capabilities \cite{bentejac2021comparative}. It is made up of a series of weak learners (typically decision trees) sequentially, with each new model attempting to correct the errors made by the previous ones and uses a gradient-based optimization algorithm to construct the trees, with the objective of minimizing a user-defined loss function.

% Advantages of XGBoost
XGBoost is one of the most efficient models in terms of accuracy and it can be applied over datasets where both categorical and numeric features are included. It often outperforms RF in terms of predictive accuracy, especially when dealing with structured/tabular data.
While RF has some mechanism to mitigate overfitting, XGBoost incorporates regularization, which provides better control.

% Disadvantages of XGBoost
On the other hand, the interpretability of the resulting models is far more complicated than a single tree and it requires specific knowledge to be able to tune its parameters. While both XGBoost and RF are ensemble models, RF models are often considered more interpretable than XGBoost due to their simpler structure.

%-------------------------------------------------------------------------------
\subsection{Explainability models}
%-------------------------------------------------------------------------------

Explainability models or techniques are used to enhance the interpretability and understanding of ML models, helping the users to understand and interpret the decisions made by these models. 

% Feature Importance Analysis
In this work we focused on using \textit{Feature Importance} \cite{saarela2021comparison} as an explainability method. Feature importance refers to a class of techniques for assigning scores to input features of a predictive model that indicates the relative importance of each feature when making a prediction.

We have included two model-specific explainability methods for tree-based models and two model-agnostic explainability methods, which we will briefly explain in the next sections.

%...............................................................................
\subsubsection{Model-specific methods}\label{sec:feature-importance}
%...............................................................................

Feature importance scores for tree-based models can be broadly split into two categories \cite{zhou2021unbiased}: 
\begin{itemize}
    \item \textbf{Split-improvement scores} that are specific to tree-based methods and that  naturally aggregate the improvement associated with each node split and can be readily recorded within the tree building process.
    \item \textbf{Permutation methods} that rely on measuring the change in value or accuracy when the values of one feature are replaced by uninformative noise, often generated by a permutation. 
\end{itemize}

%...............................................................................
\paragraph{Mean Decrease in Impurity (MDI)}

Mean Decrease in Impurity (MDI) is a split-improvement score that measures the inherent importance of different measures \cite{breiman2001random}. This is determined by how much each feature contributes to reducing the uncertainty in the target variable. MDI is measured by the amount of reduction in the Gini impurity or entropy that is achieved by splitting on a particular feature.

%...............................................................................
\paragraph{Mean Decrease Accuracy (MDA)}

Mean Decrease Accuracy (MDA) (also known as ``Permutation Importance'') \cite{breiman2001random} measures how the accuracy score decreases when a feature is not available. It randomly discards each feature and computes the change in the performance of the model, and orders them based on their impact.

%...............................................................................
\subsubsection{Model-agnostic Methods}\label{sec:model-agnostic-explicability}
%...............................................................................

As commented, model-agnostic explainability methods are independent of the specific ML model used and can be applied to any model, regardless of its architecture or underlying algorithms.

%...............................................................................
\paragraph{SHapley Additive exPlanations (SHAP)}

Presented by Lundberg and Lee \cite{lundberg2017unified}, SHAP (SHapley Additive exPlanations) is a unified framework for interpreting predictions. It is model-agnostic method that assigns each feature an importance value for a particular prediction. Its origins are related to the cooperative game theory where players collaborate together to obtain a certain score. In the AI scenario, the players are the features and the score is the prediction. 

At a general level, SHAP gives us the insights about the weight of each of the features involved in the classification process. The SHAP framework allows for a practical visualization of the Shapley value using an ``importance'' diagram. %, as figure \ref{fig:rf_shap_chart_minimum} shows. 

We are then interested in the contribution of each of the features to the final decision, which is calculated by the SHAP value. 

%...............................................................................
\paragraph{Locally Interpretable Model-agnostic Explanations (LIME)}

Proposed by Ribeiro et. al \cite{ribeiro2016why}, Locally Interpretable Model-agnostic Explanations (LIME) tries to understand the features that influence a prediction on a single instance. It focuses on a local level where a linear model is enough to explain the model behavior. LIME creates a surrogate model around the example for which we try to understand the prediction made. 

Its scope is inherently local, therefore, in order to compare the results of LIME at a global level, it is necessary to iterate over all the examples and gather the most frequent features in terms of importance.

%%%%%%%%%%%%%%%%%%%%%%%%%%%%%%%%%%%%%%%%%%%%%%%%%%%%%%%%%%%%%%%%%%%%%%%%%%%%%%%%
\section{Results}
\label{sec:results}
%%%%%%%%%%%%%%%%%%%%%%%%%%%%%%%%%%%%%%%%%%%%%%%%%%%%%%%%%%%%%%%%%%%%%%%%%%%%%%%%

In this section we will present the results of the different ML models in terms of accuracy and interpretability given the three feature sets discussed in section \ref{sec:feature_selection} and compare them with the medical guidelines and the medical criteria used as reference.

%-------------------------------------------------------------------------------
\subsection{Accuracy of the resulting models}
%-------------------------------------------------------------------------------

As the basis for the evaluation of accuracy of the different models included in this experiment we have used the following metrics:
\begin{itemize}
    \item \textbf{Accuracy} that shows how often a classification ML model is correct overall. 
    \item \textbf{Precision} that shows how often an ML model is correct when predicting the target class. 
    \item \textbf{Recall} that shows whether an ML model can find all objects of the target class. 
    \item \textbf{F1 Score} that provides a balance between precision and recall, making it a more comprehensive metric for evaluating classification models.
\end{itemize} 

In our specific scenario with a dataset unbalanced to chemotherapy cases, the F1-score provides a more robust metric. The model parameters (e.g., max depth, min samples per leaf, etc.) have been established by selecting the optimal configuration in every case. Each model was evaluated using different configurations, and we did only keep the one with higher accuracy. Results can be seen in Table \ref{tab:modelAccuracyEvaluation}.

\begin{table}[htbp!]
\footnotesize
\caption{Accuracy results for three ML models using three different feature sets.}
\label{tab:modelAccuracyEvaluation}
\begin{tabular}{@{}lllrrrrr@{}}
\toprule
Model & Parameters\footnotemark[1]\footnotemark[2] & Feature Set & Accuracy & Precision & Recall & F1-Score \\
\midrule
DT & md=3, msl=5  & Minimum        & 0,66 & 0,72 & 0,66 & 0,63 \\
DT & md=2, msl=30 & Recommended    & 0,57 & 0,58 & 0,57 & 0,55 \\
DT & md=3, msl=20 & Maximum        & 0,62 & 0,62 & 0,62 & 0,62 \\
\midrule
RF & md=2, msl=5  & Minimum        & 0,54 & 0,76 & 0,54 & 0,43 \\
RF & md=3, msl=5  & Recommended    & 0,51 & 0,76 & 0,51 & 0,38 \\
RF & md=4, msl=5  & Maximum        & 0,51 & 0,76 & 0,51 & 0,38 \\
\midrule
XGB & md=2, ne=50 & Minimum        & 0,66 & 0,70 & 0,66 & 0,64 \\
XGB & md=4, ne=40 & Recommended    & 0,54 & 0,53 & 0,54 & 0,52 \\
XGB & md=3, ne=40 & Maximum        & 0,59 & 0,62 & 0,59 & 0,55 \\
\bottomrule
\end{tabular}
\footnotetext[1]{Parameters. md: max depth, msl: min samples per leaf, ne: number of estimators.}
\footnotetext[2]{The criterion for the tree node evaluation for both the DT and RF was the Gini index. The XGBoost objective was set to binary logistic.}
\end{table}

From Table \ref{tab:modelAccuracyEvaluation} we can state that the best models in terms of accuracy have been those models built with the \textit{minimum} set of features. This circumstance suggest that adding additional information to a dataset does not always result in better accuracy, and moreover it makes the models more complex.

Referring to the type of the models, DTs are more accurate in general but their results are very close to those produced by the XGBoost. The intuition we had when using an advanced type of model, such as XGBoost, was that the increased complexity would translate into improved performance. The fact of not having any gain in accuracy could be explained by the small size of the training set, as it is known that typically ML algorithms require a big amount of data to produce acceptable results.

%-------------------------------------------------------------------------------
\subsection{Interpretability of the resulting models}\label{sec:interpretability}
%-------------------------------------------------------------------------------

Presenting interpretability information is a bit complex, since there is no defined measure of how correct the interpretation of an ML model is. We intend to move in this direction by comparing the interpretability of the models with what the medical criteria of the field expressed through guidelines and the opinion of the physicians. In the following sections we will discuss how to analyze the importance of the different features according to the different explainability methods used for the different feature sets. 

%...............................................................................
\subsubsection{Extracting the feature importance from XAI methods}\label{sec:feature_importance_XAI}
%...............................................................................

In this section we will illustrate how to extract the feature importance that the XAI methods assign to the different features given a ML model. To this end we chose a DT model (since it is transparent regarding interpretability) trained with the \textit{minimum} set of features (to make the figures more readable). The DT parameters that obtain the best accuracy are 3 as the \textit{maximum depth} of the tree, 5 as the \textit{minimum samples per leaf}, and Gini as the \textit{split criterion}. The best accuracy obtained was approximately 0.66.

\begin{figure}[htbp!]
\centering
\includegraphics[width=0.80\textwidth]{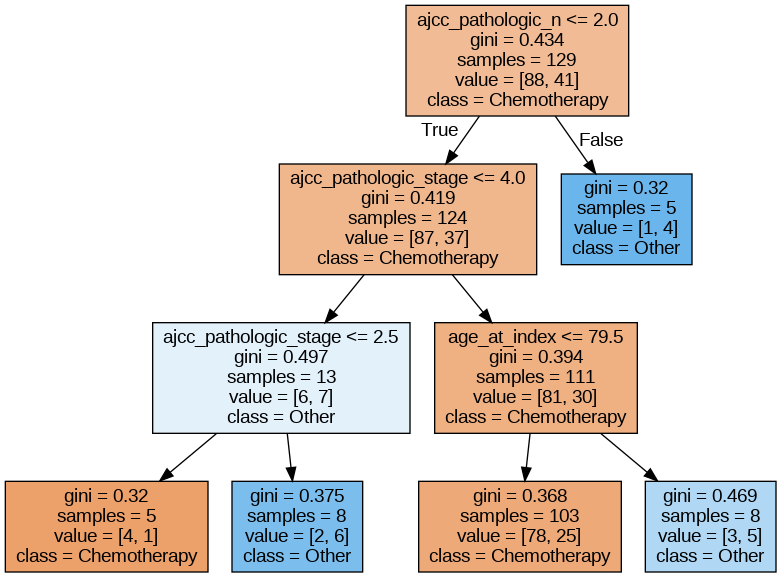}
\caption{Decision tree for the minimum set of features.}
\label{fig:dt-minimum}
\end{figure}

Figure \ref{fig:dt-minimum} shows the DT obtained. In this figure we can observe that \textit{pathologic N}, \textit{pathologic stage} and \textit{age} are the most relevant features. In the following sections we will apply different explainability methods and extract the feature importance delivered by them.

\begin{figure}[htbp!]
\centering
\includegraphics[width=0.6\textwidth]{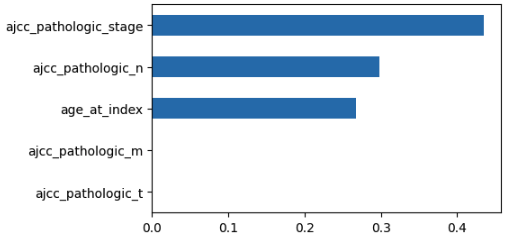}
\caption{MDI for the DT with the minimum set of features.}
\label{fig:dt-minimum-mdi}
\end{figure}

%...............................................................................
Using MDI to examine the feature importance of the classifier using the minimum feature set, we obtain the results of Figure \ref{fig:dt-minimum-mdi}. The size of each bar is related to the relevance of the feature. So, in this case, \textit{stage} is the most relevant feature followed by \textit{pathologic N} and \textit{age}.

\begin{figure}[htbp!]
\centering
\includegraphics[width=0.4\textwidth]{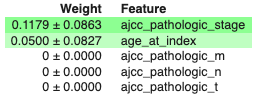}
\caption{MDA for the DT with the minimum set of features.}
\label{fig:dt-minimum-mda}
\end{figure}

%...............................................................................
On the other hand, using MDA to examine the feature importance of the classifier using the minimum feature set, we obtain the results of Figure \ref{fig:dt-minimum-mda}. Here, relevant features are highlighted. In this case, \textit{stage} is the most relevant feature followed by \textit{age}.

\begin{figure}[htbp!]
\centering
\includegraphics[width=0.8\textwidth]{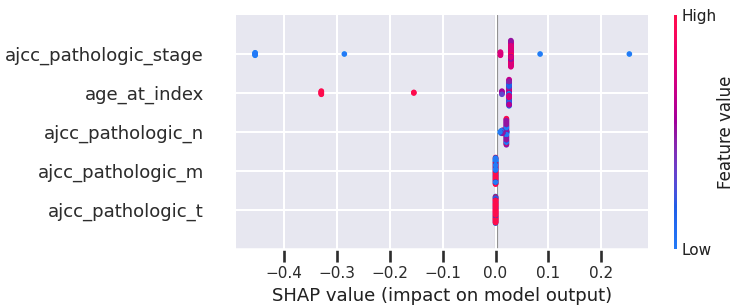}
\caption{SHAP for the DT with the minimum set of features.}
\label{fig:dt-minimum-shap}
\end{figure}

%...............................................................................
Taking into account the model-agnostic method SHAP, Figure  \ref{fig:dt-minimum-shap} presents the feature importance of the classifier using the minimum feature set. This figure is the summary chart of SHAP which presents in the left axis the features in order of relevance. In this case, \textit{stage}, \textit{age} and \textit{pathologic N} the most important ones.

\begin{figure}[htbp!]
\centering
\includegraphics[width=0.8\textwidth]{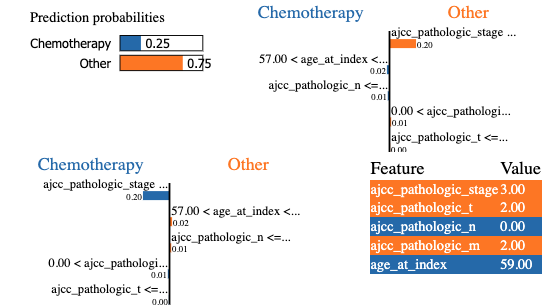}
\caption{LIME for the DT with the minimum set of features for the different ML models and XAI methods.}
\label{fig:dt-minimum-lime}
\end{figure}

%...............................................................................
Finally, LIME was applied to the same configuration. In Figure \ref{fig:dt-minimum-lime} we show the local explanation of a particular patient that indicates \textit{stage}, \textit{pathologic T} and \textit{pathologic N} as the most important features. Nevertheless, to be able to evaluate and compare LIME results at a global level we iterate over all the local explanations and keep the most frequent features for all the examples. \textit{Stage}, \textit{pathologic N} and \textit{age} are the most relevant features at a global level.

%...............................................................................

All these results for the DT classifier using the \textit{minimum} feature set are summarized in Table \ref{tab:xai_table_minimum} assigning a ``1'' to the highly relevant features, ``2'' for relevant features and ``3'' for the less relevant. Leaving blank values if the feature is not relevant for the XAI method. We also include for comparison the relevance of the features using the medical guidelines and the medical expert criteria.

\begin{table}[htbp!]
\footnotesize
\caption{Feature importance for the \textit{minimum} set of features.}
\label{tab:xai_table_minimum}
\begin{tabular}{p{0.75cm} 
                p{0.4cm}p{0.4cm}p{0.4cm}p{0.55cm} 
                p{0.4cm}p{0.4cm}p{0.4cm}p{0.55cm} 
                p{0.4cm}p{0.4cm}p{0.4cm}p{0.55cm} 
                p{0.4cm} p{0.4cm}}
\toprule
        & \multicolumn{4}{c}{DT}       & \multicolumn{4}{c}{RF}  & \multicolumn{4}{c}{XGB}   &       &      \\ 
\cmidrule(lr){1-1}\cmidrule(lr){2-5}\cmidrule(lr){6-9}\cmidrule(lr){10-13}\cmidrule(lr){14-15}
Feature & MDI & MDA & SHAP & LIME      & MDI & MDA & SHAP & LIME & MDI & MDA & SHAP & LIME   & Guid. & Exp. \\
\cmidrule(lr){1-1}\cmidrule(lr){2-5}\cmidrule(lr){6-9}\cmidrule(lr){10-13}\cmidrule(lr){14-15}     
Age     & 2   & 2   & 1    & 3         & 1   & 3   & 2    & 3    & 2   & 1   & 1    & 2      & 3     & 2    \\
Stage   & 1   & 1   & 1    & 1         & 1   & 1   & 1    & 1    & 1   & 1   & 1    & 1      & 1     & 1    \\
T       &     &     &      &           & 3   & 3   & 3    &      & 2   & 3   & 2    & 2      & 2     & 3    \\
N       & 2   &     & 3    & 3         & 2   & 1   & 2    & 1    & 2   & 2   & 3    &        & 2     & 2    \\
M       &     &     &      &           & 3   &     & 2    & 3    & 2   & 2   & 2    & 2      & 2     & 1    \\
\bottomrule
\end{tabular}
\end{table}

%...............................................................................
\subsubsection{Similarity index}
%...............................................................................

We need a metric that allows us to compare the results obtained. One good candidate is the \textit{Jaccard Similarity index} \cite{haeupler2014consistent} that measures the similarity between two sets dividing the size of the intersection by the size of the union of the two sets. Mathematically, given two finite sets $S$ and $T$ from a universe $U$, the  \textit{Jaccard similarity} of $S$ and $T$ is defined as:
\begin{align*}
jacc(S,T) = \frac{|S \cap T|}{|S \cup T|}
\end{align*}

In our case the elements of the set are the different features and the importance that the different models of explainability give to these features. In addition, we not only want to see if the elements of the different sets coincide, we also want to see if they coincide in order, giving more importance to the coincidences in the first positions. For that reason the \textit{Weighted Jaccard Similarity index} \cite{haeupler2014consistent} is more convenient since it allows to handle cases where elements in the data have different weights or importance.

The Weighted Jaccard Similarity index assumes that a weighted set exists that associates a real weight to each element in it (being this weight $\geq 0$). Therefore a weighted set is defined by a map $w : U \rightarrow \mathbb{R}$. %, with the weight of the elements outside the set defined as 0.
In our case, since we want to give more importance to the first position matches we assign a weight of $3$ to the first position, $2$ to the second position, $1$ for the third position and $0$ for the non-ranked elements.

Thus, given two vectors of weights $\mathbf{x} = (x_1, x_2, \ldots, x_n)$ and $\mathbf{y} = (y_1, y_2, \ldots, y_n)$ with all real $x_i, y_i \geq 0$, then their weighted Jaccard similarity coefficient is defined as:

\begin{align*}
J_\mathcal{W}(\mathbf{x}, \mathbf{y}) = \frac{\sum_i \min(x_i, y_i)}{\sum_i \max(x_i, y_i)}
\end{align*}

In Table \ref{tab:xai_similarity_minimum_DT} and Table \ref{tab:xai_similarity_minimum_XGB} we can see the results of calculating the weighted Jaccard index for the DT and XGB models respectively according to the data included in Table \ref{tab:xai_table_minimum}. 
We decided to focus our analysis in the two models that offer the best performance. 

\begin{table}[htbp!]
\footnotesize
\caption{Similarity measures for the different XAI methods using the DT model and the \textit{minimum} set of features.}
\label{tab:xai_similarity_minimum_DT}
\begin{tabular}{ccccccc}
\toprule
            & MDI-DT            & MDA-DT            & SHAP-DT            & LIME-DT          & Guidelines      & Experts         \\
\cmidrule(lr){1-1}\cmidrule(lr){2-5}\cmidrule(lr){6-7}
MDI-DT      & \cellcolor[gray]
              {0.8}1.00         & 0.71              & 0.75               & 0.71             & 0.55            & 0.64            \\
MDA-DT      & 0.71              & \cellcolor[gray]
                                  {0.8}1.00         & 0.71               & 0.67             & 0.36            & 0.45            \\
SHAP-DT     & 0.75              & 0.71              & \cellcolor[gray]
                                                              {0.8}1.00  & 0.71             & 0.42            & 0.50            \\
LIME-DT     & 0.71              & 0.67              & 0.71               & \cellcolor[gray]
                                                                           {0.8}1.00        & 0.50            & 0.45            \\
\midrule                                                                           
Guidelines  & 0.55              & 0.36              & 0.42               & 0.50             & \cellcolor[gray]
                                                                                              {0.8}1.00       & 0.75            \\
Experts     & 0.64              & 0.45              & 0.50               & 0.45             & 0.75            & \cellcolor[gray]
                                                                                                                {0.8}1.00       \\
\bottomrule
\end{tabular}
\end{table}

\begin{table}[htbp!]
\footnotesize
\caption{Similarity measures for the different XAI methods using the XGB model and the \textit{minimum} set of features.}
\label{tab:xai_similarity_minimum_XGB}
\begin{tabular}{ccccccc}
\toprule
            & MDI-XGB            & MDA-XGB            & SHAP-XGB            & LIME-XGB          & Guidelines      & Experts         \\
\cmidrule(lr){1-1}\cmidrule(lr){2-5}\cmidrule(lr){6-7}
MDI-XGB      & \cellcolor[gray]
              {0.8}1.00         & 0.83              & 0.83               & 0.81             & 0.91            & 0.83            \\
MDA-XGB      & 0.83              & \cellcolor[gray]
                                  {0.8}1.00         & 0.83               & 0.67             & 0.75            & 0.83            \\
SHAP-XGB     & 0.83              & 0.83              & \cellcolor[gray]
                                                              {0.8}1.00  & 0.82             & 0.75            & 0.69            \\
LIME-XGB     & 0.81              & 0.67              & 0.82               & \cellcolor[gray]
                                                                           {0.8}1.00        & 0.73            & 0.67            \\
\midrule                                                                           
Guidelines  & 0.91              & 0.75              & 0.75               & 0.73             & \cellcolor[gray]
                                                                                              {0.8}1.00       & 0.75            \\
Experts     & 0.83              & 0.83              & 0.69               & 0.67             & 0.75            & \cellcolor[gray]
                                                                                                                {0.8}1.00       \\
\bottomrule
\end{tabular}
\end{table}

From these tables we can see several things, firstly that the degree of similarity between the guides and the experts is high (0.75), and the same occurs between the different methods of explainability. We see that the main similarity is that they all agree in assigning the \textit{stage} as the most important feature, while the main difference is that the ML models give less importance to features such as \textit{T} and \textit{M}.

%...............................................................................
\subsubsection{Feature importance using the \textit{recommended} feature set}\label{sec:xai_recommended}
%...............................................................................

In this section we will present the interpretability results for each ML model and each feature relevance method using the \textit{recommended} features set. The results are summarized in Table \ref{tab:xai_table_recommended}.

\begin{table}[htbp!]
\footnotesize
\caption{Feature importance for the \textit{recommended} set of features.}
\label{tab:xai_table_recommended}
\begin{tabular}{p{1cm}  p{0.3cm}p{0.3cm}p{0.4cm}p{0.55cm}p{0.3cm}p{0.3cm}p{0.4cm}p{0.55cm}p{0.3cm}p{0.3cm}p{0.4cm}p{0.55cm}p{0.4cm} p{0.4cm}}
\toprule
            & \multicolumn{4}{c}{DT}    & \multicolumn{4}{c}{RF}    & \multicolumn{4}{c}{XGB} & &  \\ 
\cmidrule(lr){1-1}\cmidrule(lr){2-5}\cmidrule(lr){6-9}\cmidrule(lr){10-13}\cmidrule(lr){14-15}
Feature     & MDI & MDA & SHAP & LIME   & MDI & MDA & SHAP & LIME   & MDI & MDA & SHAP & LIME & Guid. & Exp. \\
\cmidrule(lr){1-1}\cmidrule(lr){2-5}\cmidrule(lr){6-9}\cmidrule(lr){10-13}\cmidrule(lr){14-15}
Age         & 1   & 2   & 1    &        & 1   & 3   & 2    &        & 1   & 2   & 1    & 2    & 3     & 2 \\
Stage       & 2   & 1   & 2    & 1      & 1   & 1   & 1    & 1      & 1   & 2   & 2    & 1    & 1     & 1 \\
N           &     &     &      &        & 2   & 1   & 2    & 3      & 2   &     & 3    &      & 2     & 2 \\
M           &     &     &      &        & 2   & 2   & 3    &        & 3   &     & 3    &      & 2     & 1 \\ 
\midrule
Year        & 2   & 1   & 1    & 2      & 1   & 1   & 1    & 3      & 2   &     &      & 1    & 3     & 2 \\
Adeno.      &     &     &      &        &     &     &      &        &     &     &      &      & 2     & 2 \\
Type        &     &     &      &        & 2   & 1   & 2    &        & 1   & 1   & 3    &      & 2     & 1 \\
Status      & 2   & 2   & 1    & 1      & 3   & 1   & 3    &        & 1   & 2   & 2    &      & 1     & 1 \\
Grade       &     &     &      &        & 2   & 2   & 2    &        & 2   & 1   & 3    &      & 3     & 3 \\
Dimensi.    & 3   &     & 3    &        & 2   & 1   & 3    &        & 2   &     & 2    &      & 1     & 1 \\
Residual    & 1   &     & 2    & 2      & 2   &     & 1    & 1      & 2   &     & 1    & 1    & 3     & 1 \\
Diagnos.    &     &     &      &        & 3   & 1   & 3    &        & 1   & 1   & 2    &      & 2     & 2 \\
Surgery     &     &     &      &        & 3   & 1   & 2    &        & 3   &     &      &      & 2     & 2 \\
Lymph       &     &     &      &        & 2   & 1   & 2    &        & 2   &     & 2    &      & 1     & 1 \\ 
\bottomrule
\end{tabular}
\end{table}

In Table \ref{tab:xai_similarity_recommended_DT} and Table \ref{tab:xai_similarity_recommended_XGB} we can see the similarity results for the DT and XGB models respectively according to the data included in Table \ref{tab:xai_table_recommended}. From these two tables we can see that again there is a high similarity between experts and guides, and that the similarity between the XAI methods is quite consistent (except for LIME). 

\begin{table}[htbp!]
\footnotesize
\caption{Similarity measures for the different XAI methods using the DT model and the \textit{recommended} set of features.}
\label{tab:xai_similarity_recommended_DT}
\begin{tabular}{ccccccc}
\toprule
            & MDI-DT            & MDA-DT            & SHAP-DT            & LIME-DT          & Guidelines      & Experts         \\
\cmidrule(lr){1-1}\cmidrule(lr){2-5}\cmidrule(lr){6-7}
MDI-DT      & \cellcolor[gray]
              {0.8}1.00         & 0.53              & 0.80               & 0.53             & 0.24            & 0.34            \\
MDA-DT      & 0.53              & \cellcolor[gray]
                                  {0.8}1.00         & 0.60               & 0.54             & 0.23            & 0.26            \\
SHAP-DT     & 0.80              & 0.60              & \cellcolor[gray]
                                                              {0.8}1.00  & 0.60             & 0.27            & 0.33            \\
LIME-DT     & 0.53              & 0.54              & 0.60               & \cellcolor[gray]
                                                                           {0.8}1.00        & 0.27            & 0.29            \\
\midrule
Guidelines  & 0.24              & 0.23              & 0.27               & 0.27             & \cellcolor[gray]
                                                                                              {0.8}1.00       & 0.82            \\
Experts     & 0.34              & 0.26              & 0.33               & 0.29             & 0.82            & \cellcolor[gray]
                                                                                                                {0.8}1.00       \\
\bottomrule
\end{tabular}
\end{table}

\begin{table}[htbp!]
\footnotesize
\caption{Similarity measures for the different XAI methods using the XGB model and the \textit{recommended} set of features.}
\label{tab:xai_similarity_recommended_XGB}
\begin{tabular}{ccccccc}
\toprule
            & MDI-XGB            & MDA-XGB            & SHAP-XGB            & LIME-XGB          & Guidelines      & Experts         \\
\cmidrule(lr){1-1}\cmidrule(lr){2-5}\cmidrule(lr){6-7}
MDI-XGB      & \cellcolor[gray]
              {0.8}1.00         & 0.47              & 0.68               & 0.28             & 0.63            & 0.70            \\
MDA-XGB      & 0.47              & \cellcolor[gray]
                                  {0.8}1.00         & 0.35               & 0.21             & 0.30            & 0.32            \\
SHAP-XGB     & 0.68              & 0.35              & \cellcolor[gray]
                                                              {0.8}1.00  & 0.29             & 0.50            & 0.58            \\
LIME-XGB     & 0.28              & 0.21              & 0.29               & \cellcolor[gray]
                                                                           {0.8}1.00        & 0.20            & 0.24            \\
\midrule
Guidelines  & 0.63              & 0.30              & 0.50               & 0.20             & \cellcolor[gray]
                                                                                              {0.8}1.00       & 0.82            \\
Experts     & 0.70              & 0.32              & 0.58               & 0.24             & 0.82            & \cellcolor[gray]
                                                                                                                {0.8}1.00       \\
\bottomrule
\end{tabular}
\end{table}

If we focus on the DT data we can see that the decision trees take into consideration fewer features than the experts and the guides, which is normal since they only consider those that are necessary to establish a decision. This results in low similarity values. On the other hand the XGB model spreads the importance along the different measures, therefore the similarity values with experts and guides are higher. 
we see that the features chosen as the important ones by the DT model and according to the XAI methods are: \textit{Age}, \textit{Stage}, \textit{Year of initial diagnosis}, \textit{Neoplasm cancer status} and \textit{Residual tumor}. Except for \textit{Age} and \textit{Year} the rest are considered very or fairly important by experts and guides. On the other hand, XGB considers again features such as \textit{Age}, \textit{Stage}, \textit{Residual tumor} or \textit{Neoplasm cancer status}. But the biggest difference is that it considers features that are completely irrelevant for the DTs but relevant for experts and guides such as \textit{Histological type} or \textit{Initial diagnosis method}.

%...............................................................................
\subsubsection{Feature importance using the \textit{maximum} feature set}\label{sec:xai_maximum}
%...............................................................................

In this section the interpretability results for each ML model and each features relevance method using the \textit{maximum} features set is presented. Table \ref{tab:xai_table_maximum} summarize the results. 

\begin{table}[htbp!]
\footnotesize
\caption{Feature importance for the \textit{maximum} set of features.}
\label{tab:xai_table_maximum}
\begin{tabular}{p{1cm}  p{0.3cm}p{0.3cm}p{0.4cm}p{0.55cm}p{0.3cm}p{0.3cm}p{0.4cm}p{0.55cm}p{0.3cm}p{0.3cm}p{0.4cm}p{0.55cm}p{0.4cm} p{0.4cm}}
\toprule
            & \multicolumn{4}{c}{DT}    & \multicolumn{4}{c}{RF}    & \multicolumn{4}{c}{XGB} & &  \\ 
\cmidrule(lr){1-1}\cmidrule(lr){2-5}\cmidrule(lr){6-9}\cmidrule(lr){10-13}\cmidrule(lr){14-15}
Feature     & MDI & MDA & SHAP & LIME   & MDI & MDA & SHAP & LIME   & MDI & MDA & SHAP & LIME & Guid. & Exp. \\
\cmidrule(lr){1-1}\cmidrule(lr){2-5}\cmidrule(lr){6-9}\cmidrule(lr){10-13}\cmidrule(lr){14-15}
Age         & 3   & 2   & 3    &        & 1   &     & 2    & 2      & 1   &     & 1    & 2      & 3     & 2 \\
Stage       & 2   & 2   & 1    & 2      & 1   & 1   & 1    & 1      & 2   & 1   & 1    & 1      & 1     & 1 \\
T           &     &     &      &        & 2   & 2   & 3    &        &     &     &      &        & 2     & 3 \\
N           &     &     &      &        & 2   & 1   & 2    & 3      & 3   &     &      &        & 2     & 2 \\
M           &     &     &      &        & 3   & 2   & 3    &        & 3   &     &      &        & 2     & 1 \\ 
\midrule
Year        & 1   & 1   & 1    &        & 1   & 3   & 1    & 3      & 2   &     & 1    & 3      & 3     & 1 \\
Adeno.      &     &     &      &        &     &     & 3    &        &     &     &      &        & 2     & 1 \\
Type        &     &     &      & 3      & 2   & 1   & 3    &        & 3   & 1   & 2    &        & 2     & 1 \\
Status      &     &     &      &        & 3   &     & 3    &        & 3   &     & 2    &        & 1     & 2 \\
Grade       &     &     &      &        & 2   & 2   & 3    &        & 3   & 2   & 3    &        & 3     & 1 \\
Dimensi.    &     &     &      &        & 1   &     & 3    &        & 1   & 3   & 2    &        & 1     & 1 \\
Residual    &     &     &      &        & 2   &     & 1    & 2      & 3   & 2   & 1    &        & 3     & 1 \\
Diagnos.    &     &     &      &        & 3   &     & 3    &        & 3   &     & 2    &        & 2     & 2 \\
Surgery     &     &     &      &        & 3   &     & 3    &        &     &     &      &        & 2     & 2 \\
Lymph       &     &     &      &        & 2   & 3   & 3    &        & 2   &     & 2    &        & 1     & 1 \\  
\midrule
Gender      &     &     &      &        & 2   &     &      &        & 2   & 1   & 1    &        &       & \\
Race        &     &     &      &        & 2   &     & 2    & 2      & 3   &     & 3    & 3      &       & \\
Ethnic.     &     &     &      &        & 3   &     & 3    &        &     &     & 3    &        &       & \\
Other DX    &     &     &      &        & 3   &     & 3    &        &     &     &      &        &       & \\
Diabetes    &     &     &      &        & 2   &     & 3    &        & 3   &     &      &        & 3     & 3 \\
Family      & 2   &     & 1    & 1      & 2   & 2   & 2    &        & 3   & 1   & 1    & 1      & 2     & 3 \\
Radiat.     &     &     &      &        & 2   & 3   & 3    &        & 3   & 2   & 3    &        & 2     & 2 \\
Therapy     &     &     &      &        & 3   &     & 2    &        & 3   &     & 3    &        & 2     & 1 \\
N.tumor     &     &     &      &        & 3   &     & 3    &        &     & 2   &      &        &       & 1 \\
Days to     &     &     &      &        & 2   &     &      &        & 3   &     &      &        &       & 1 \\
Tobacco     &     &     &      &        & 2   & 3   & 3    &        & 2   &     & 2    &        &       & \\
Alcohol     &     &     &      &        & 2   &     &      &        & 3   &     & 2    &        &       & \\
\bottomrule
\end{tabular}
\end{table}

In Table \ref{tab:xai_similarity_maximum_DT} and Table \ref{tab:xai_similarity_maximum_XGB} we can see the similarity results for the DT and XGB models respectively according to the data included in Table \ref{tab:xai_table_maximum}. As there are more features, the similarity values decrease, which is to some extent normal since the ambiguity inherent in assigning concrete values of importance to the different measures must be taken into account. 

\begin{table}[htbp!]
\footnotesize
\caption{Similarity measures for the different XAI methods using the DT model and the \textit{maximum} set of features.}
\label{tab:xai_similarity_maximum_DT}
\begin{tabular}{ccccccc}
\toprule
            & MDI-DT            & MDA-DT            & SHAP-DT            & LIME-DT          & Guidelines      & Experts         \\
\cmidrule(lr){1-1}\cmidrule(lr){2-5}\cmidrule(lr){6-7}
MDI-DT      & \cellcolor[gray]
              {0.8}1.00         & 0.67              & 0.80               & 0.40             & 0.15            & 0.13            \\
MDA-DT      & 0.67              & \cellcolor[gray]
                                  {0.8}1.00         & 0.55               & 0.18             & 0.10            & 0.14            \\
SHAP-DT     & 0.80              & 0.55              & \cellcolor[gray]
                                                              {0.8}1.00  & 0.45             & 0.18            & 0.15            \\
LIME-DT     & 0.40              & 0.18              & 0.45               & \cellcolor[gray]
                                                                           {0.8}1.00        & 0.13            & 0.08            \\
\midrule
Guidelines  & 0.15              & 0.10              & 0.18               & 0.13             & \cellcolor[gray]
                                                                                              {0.8}1.00       & 0.63            \\
Experts     & 0.13              & 0.14              & 0.15               & 0.08             & 0.63            & \cellcolor[gray]
                                                                                                                {0.8}1.00       \\
\bottomrule
\end{tabular}
\end{table}

\begin{table}[htbp!]
\footnotesize
\caption{Similarity measures for the different XAI methods using the XGB model and the \textit{maximum} set of features.}
\label{tab:xai_similarity_maximum_XGB}
\begin{tabular}{ccccccc}
\toprule
            & MDI-XGB            & MDA-XGB            & SHAP-XGB            & LIME-XGB          & Guidelines      & Experts         \\
\cmidrule(lr){1-1}\cmidrule(lr){2-5}\cmidrule(lr){6-7}
MDI-XGB      & \cellcolor[gray]
              {0.8}1.00         & 0.24              & 0.60               & 0.26             & 0.43            & 0.40            \\
MDA-XGB      & 0.24              & \cellcolor[gray]
                                  {0.8}1.00         & 0.38               & 0.26             & 0.26            & 0.29            \\
SHAP-XGB     & 0.60              & 0.38              & \cellcolor[gray]
                                                              {0.8}1.00  & 0.35             & 0.40            & 0.40            \\
LIME-XGB     & 0.26              & 0.26              & 0.35               & \cellcolor[gray]
                                                                           {0.8}1.00        & 0.19            & 0.18            \\
\midrule
Guidelines  & 0.43              & 0.26              & 0.40               & 0.19             & \cellcolor[gray]
                                                                                              {0.8}1.00       & 0.63            \\
Experts     & 0.40              & 0.29              & 0.40               & 0.18             & 0.63            & \cellcolor[gray]
                                                                                                                {0.8}1.00       \\
\bottomrule
\end{tabular}
\end{table}

From these tables we can see that in DT the results of the XAI methods are quite consistent with each other, highlighting few features but almost always the same (\textit{Age}, \textit{Stage}, \textit{Year}, \textit{Family history}). XGB takes into account the influence of more features but the explanatory methods are not as consistent with each other, indicating that it is a more opaque model from which it is more difficult to find consistent explanatory measures.

%%%%%%%%%%%%%%%%%%%%%%%%%%%%%%%%%%%%%%%%%%%%%%%%%%%%%%%%%%%%%%%%%%%%%%%%%%%%%%%%
\section{Discussion }
\label{sec:discussion}
%%%%%%%%%%%%%%%%%%%%%%%%%%%%%%%%%%%%%%%%%%%%%%%%%%%%%%%%%%%%%%%%%%%%%%%%%%%%%%%%

%...............................................................................
\paragraph{Performance}

% Results in accuracy
Regarding the performance of the different ML models, as described throughout the results section, the accuracy obtained is low; all of them present similar accuracy values near, but not reaching 0.7. There is a justification for such low precision values and, in any case, they are not entirely relevant to the objective we are pursuing in the paper. First of all, we can say that the dataset has few cases, so ML models suffer when trying to generalize patterns present in the data. This is a clear \textit{data bottleneck} problem and, as discussed in \cite{mosqueira2024addressing}, a possible solution is to use \textit{data augmentation} strategies to improve data quality and quantity. In that work the accuracy increased more than 10 percent with the collaboration of human experts who helped to improve the labeling and the generation of synthetic cases.

However, the aim of this paper was to focus on how to evaluate the explanatory capabilities of ML models and to analyze the consistency between different XAI methods. Adding synthetic data could improve accuracy but at the cost of introducing correlations that might not be real and could affect explanatory capabilities.

% ML models comparison
Comparing the different ML models, we can see that DT and XGBoost models deliver similar results in terms of accuracy in all scenarios. The fact of increasing or decreasing the set of features did not make the RF model nor the XGBoost model better in terms of performance. It could be due to the small number of examples used to train the model. Normally, they are expected to find complex relationships between the different features but in our work it was not the case. Although all ML models are tree-based and that bagging and boosting techniques are included in the RF and XGBoost models respectively, we have found no gain in terms of accuracy nor in explicability matters.

% Feature sets comparison
Regarding the different features sets used, it is interesting to highlight that the \textit{minimum} feature set was the one that obtained better results in terms of accuracy. One possible explanation could be the type of features included in this set. Apart from \textit{age}, features like \textit{pathologic T} (code to define the size or contiguous extension of the primary tumor), \textit{pathologic N} (code to represent the stage of cancer based on the nodes present), \textit{pathologic M} (code to represent the defined absence or presence of distant spread or metastases to locations via vascular channels or lymphatics beyond the regional lymph nodes) and \textit{stage} (code to represent the extent of a cancer, especially whether the disease has spread from the original site to other parts of the body) are summarizing codes of the cancer status. Therefore, they are a sort of dimensionality-reduction technique applied by physicians to better understand and communicate cancer status. In these features is included much of the information needed to decide whether to give chemotherapy treatment although, obviously, the detailed treatment itself requires more information. 
The remaining sets of features offer lower results, but very close to the results of the \textit{minimum} feature set. 

%...............................................................................
\paragraph{Human-in-the-Loop}

The participation of humans in the process of developing ML systems (a.k.a., Human-in-the-loop or HITL) could help in enhancing transparency, improving trust, and achieving better performance \cite{guillot2022human}.  Nowadays, there exist many tools supporting the inclusion of human experts both as ML practitioners or as domain experts \cite{mosqueira2022classification}.

% Feature engineering with experts
Regarding Feature Engineering (FE) we have seen that it is an important part of every ML project. There have been other works that focus in HITL feature engineering process. For example, Anderson et al. \cite{anderson2016runtime} propose an interactive HITL FE scheme, based on state-of-the-art dimensionality reduction for \textit{nowcasting} features for economic and social science models; and Gkorou et al. \cite{gkorou2020get} propose an interactive FE scheme based on dimensionality reduction for Integrated Circuit (IC) manufacturing. Both cases show that by engineering features it is possible to obtain higher predictive capabilities and to improve the interpretability of the model. 

% In our case - HITL with experts
In our case we collaborate with medical experts for: 1) selecting the best features available from the specialized dataset and, 2) establishing a relevance value creating a recommended set of features and giving insights about the pancreatic cancer diagnosis. This collaboration required an active participation of human experts, something that add complexity to the process. 

% Indirect use of experts through guides
But we also have use medical expertise through medical guidelines, that are standard procedures in the field. This process of analyzing medical guidelines do not require the active participation of human experts but have also its complexities as this guidelines tend to be exhaustive and sometimes hard to navigate through. 

% Evaluating explainability methods
This collaboration with human experts of the domain and the analysis of guidelines available in the same domain can be used not only as a feature engineering process, but also as a way to evaluate explainability methods.

%...............................................................................
\paragraph{Explainability}

% XAI methods
When it comes to explainability, it is important to notice that each approach available measure specific responses of the model using different techniques, and thus they could produce different results. However, in this work we have been able to see that, as a general rule, the explanatory methods, given a given ML model, provide fairly consistent results among themselves, with high levels of similarity.

% MDI similar to SHAP
We can see that, methods such as MDI and SHAP use very different procedures to reach their conclusions, but they are very similar in terms of feature importance. Also SHAP provides several charts that allow us to better understand the explainability results and it has also the advantage of being model-agnostic, so that we can compare the results over different ML models.

% How to interpret XAI results
In any case, we should not expect these tools on their on to produce human understanding of the models. Normally, they require a thoughtful interpretation to understand the underlying decision making process. Furthermore, explainability of black-box models have been criticized because the explanations either do not provide enough detail to understand what the black-box is doing or even they do not make sense \cite{rudin2019stop}.

% Similarity as a measure to interpret XAI results
% DT vs XGB
To try to facilitate the interpretation of these explanatory results, we use a similarity measure such as the Weighted Jaccard Similarity coefficient. From their results we can see several things: Firstly, models like DT are more ``decisive'' when it comes to giving importance to features, for this ML model only a few features are important for the final result. On the other hand, and due to the very nature of the model, XGB distributes this importance more evenly among the different features. This has an effect on our results, as guides and experts also distribute the importance among the different features the similarity index of XGB with experts and guides is usually higher than that of DT.

% XGB takes more features into account
We could say then that this behavior of XGB is more ``human'', since it tries to take more information into account when making a decision. Although in this case taking into account more features does not correspond to a better performance, this may be due to problems inherent to the dataset itself and its low number of cases. Thus, for example, we see that the DT model eliminates from its consideration features such as \textit{Histological type} or \textit{Initial diagnosis} that are taken into account by experts and guides. A model such as XGB does take them into account.

% Experts and Guidelines
Regarding the importance given by the experts and the guides to the different features, we can see that the similarity index between them is always high, with some variations that are easy to explain by the subjectivity of the assignment of values. 

Regarding how the models assign importance to the features in comparison with guidelines and experts, in general we can say that the features pointed out as relevant by physicians and guidelines have been taken into account by the models but with some notable exceptions. Thus, for example, staging M has been generally forgotten by ML models when physicians assigned it a high importance. On the other hand, we have also observed that some features with low relevance for the experts, are considered by the XAI methods, such as, for example the \textit{Year of initial diagnosis}. 

%%%%%%%%%%%%%%%%%%%%%%%%%%%%%%%%%%%%%%%%%%%%%%%%%%%%%%%%%%%%%%%%%%%%%%%%%%%%%%%%
\section{Conclusions}
\label{sec:conclusions}
%%%%%%%%%%%%%%%%%%%%%%%%%%%%%%%%%%%%%%%%%%%%%%%%%%%%%%%%%%%%%%%%%%%%%%%%%%%%%%%%

% symbolic AI
As final conclusions we can highlight that, in this last ``spring'' of AI, researchers have decided to leave symbolic AI aside because its results were not entirely good and these AI models were very difficult to scale to larger problems. However, by embracing ML and subsymbolic AI we are getting better results, but these results have a worse interpretability because they have lost that symbolic character.

% Giving symbolism to what does not have symbolism
By trying to analyze ML results and compare their explainability through information from domain experts and guides what we are trying to do is to give ``symbolism'' to these results, to try to understand them from a human point of view. This approach is not problem-free, it requires active collaboration on the part of physicians, and medical guidelines are complex documents that require time for analysis. 

% Similarity measures
However, we believe that the similarity measures used to analyze the importance of the different features can be an interesting way to evaluate explainability methods. The fact that the best models are the most opaque may cause such models to be discarded in critical domains such as health. Explainability agnostic methods and similarity measures can be used as techniques to better understand how the model works and whether the model adheres to recommended medical practices.

% Feature engineering and Feature importance
Until now, the feature importance was used as a measure of dimensionality reduction, allowing to simplify the problem being modeled in order to try to speed up the learning process and, in some cases, to improve the performance of the ML model. In our case what we intend to do is to use the feature importance as a way of evaluating the explainability of the different models.

% Future work
As future work we plan to further develop the field of explainability metrics, we believe that in the future this type of metrics will be as common as performance metrics (accuracy, F1-score) are today. The idea is not only to choose the model that offers the best performance, but also the model that can best explain its conclusions and whose reasoning is best aligned with human knowledge of the domain.

%%%%%%%%%%%%%%%%%%%%%%%%%%%%%%%%%%%%%%%%%%%%%%%%%%%%%%%%%%%%%%%%%%%%%%%%%%%%%%%%
\bmhead{Acknowledgements}
%%%%%%%%%%%%%%%%%%%%%%%%%%%%%%%%%%%%%%%%%%%%%%%%%%%%%%%%%%%%%%%%%%%%%%%%%%%%%%%%

This work has been supported by the State Research Agency of the Spanish Government (Grant PID2019-107194GB-I00/AEI/10.13039/501100011033) and by the Xunta de Galicia (Grant ED431C 2022/44), supported by the EU European Regional Development Fund (ERDF). We wish to acknowledge support received from the Centro de Investigación de Galicia CITIC, funded by the Xunta de Galicia and ERDF (Grant ED431G 2019/01). The results published here are in whole or part based upon data generated by the TCGA Research Network: https://www.cancer.gov/tcga

%%%%%%%%%%%%%%%%%%%%%%%%%%%%%%%%%%%%%%%%%%%%%%%%%%%%%%%%%%%%%%%%%%%%%%%%%%%%%%%%
\bibliography{HITL-Pancreas-XAI-bibliography}% common bib file


\begin{thebibliography}{60}
\ifx \bisbn   \undefined \def \bisbn  #1{ISBN #1}\fi
\ifx \binits  \undefined \def \binits#1{#1}\fi
\ifx \bauthor  \undefined \def \bauthor#1{#1}\fi
\ifx \batitle  \undefined \def \batitle#1{#1}\fi
\ifx \bjtitle  \undefined \def \bjtitle#1{#1}\fi
\ifx \bvolume  \undefined \def \bvolume#1{\textbf{#1}}\fi
\ifx \byear  \undefined \def \byear#1{#1}\fi
\ifx \bissue  \undefined \def \bissue#1{#1}\fi
\ifx \bfpage  \undefined \def \bfpage#1{#1}\fi
\ifx \blpage  \undefined \def \blpage #1{#1}\fi
\ifx \burl  \undefined \def \burl#1{\textsf{#1}}\fi
\ifx \doiurl  \undefined \def \doiurl#1{\url{https://doi.org/#1}}\fi
\ifx \betal  \undefined \def \betal{\textit{et al.}}\fi
\ifx \binstitute  \undefined \def \binstitute#1{#1}\fi
\ifx \binstitutionaled  \undefined \def \binstitutionaled#1{#1}\fi
\ifx \bctitle  \undefined \def \bctitle#1{#1}\fi
\ifx \beditor  \undefined \def \beditor#1{#1}\fi
\ifx \bpublisher  \undefined \def \bpublisher#1{#1}\fi
\ifx \bbtitle  \undefined \def \bbtitle#1{#1}\fi
\ifx \bedition  \undefined \def \bedition#1{#1}\fi
\ifx \bseriesno  \undefined \def \bseriesno#1{#1}\fi
\ifx \blocation  \undefined \def \blocation#1{#1}\fi
\ifx \bsertitle  \undefined \def \bsertitle#1{#1}\fi
\ifx \bsnm \undefined \def \bsnm#1{#1}\fi
\ifx \bsuffix \undefined \def \bsuffix#1{#1}\fi
\ifx \bparticle \undefined \def \bparticle#1{#1}\fi
\ifx \barticle \undefined \def \barticle#1{#1}\fi
\bibcommenthead
\ifx \bconfdate \undefined \def \bconfdate #1{#1}\fi
\ifx \botherref \undefined \def \botherref #1{#1}\fi
\ifx \url \undefined \def \url#1{\textsf{#1}}\fi
\ifx \bchapter \undefined \def \bchapter#1{#1}\fi
\ifx \bbook \undefined \def \bbook#1{#1}\fi
\ifx \bcomment \undefined \def \bcomment#1{#1}\fi
\ifx \oauthor \undefined \def \oauthor#1{#1}\fi
\ifx \citeauthoryear \undefined \def \citeauthoryear#1{#1}\fi
\ifx \endbibitem  \undefined \def \endbibitem {}\fi
\ifx \bconflocation  \undefined \def \bconflocation#1{#1}\fi
\ifx \arxivurl  \undefined \def \arxivurl#1{\textsf{#1}}\fi
\csname PreBibitemsHook\endcsname

%%% 1
\bibitem[\protect\citeauthoryear{Linardatos
  et~al.}{2021}]{linardatos2021explainable}
\begin{botherref}
\oauthor{\bsnm{Linardatos}, \binits{P.}},
\oauthor{\bsnm{Papastefanopoulos}, \binits{V.}},
\oauthor{\bsnm{Kotsiantis}, \binits{S.}}:
Explainable ai: A review of machine learning interpretability methods.
Entropy
\textbf{23}(1)
(2021)
\doiurl{10.3390/e23010018}
\end{botherref}
\endbibitem

%%% 2
\bibitem[\protect\citeauthoryear{Angelov et~al.}{2021}]{angelov2021explainable}
\begin{barticle}
\bauthor{\bsnm{Angelov}, \binits{P.P.}},
\bauthor{\bsnm{Soares}, \binits{E.A.}},
\bauthor{\bsnm{Jiang}, \binits{R.}},
\bauthor{\bsnm{Arnold}, \binits{N.I.}},
\bauthor{\bsnm{Atkinson}, \binits{P.M.}}:
\batitle{Explainable artificial intelligence: an analytical review}.
\bjtitle{Wiley Interdisciplinary Reviews: Data Mining and Knowledge Discovery}
\bvolume{11}(\bissue{5}),
\bfpage{1424}
(\byear{2021})
\end{barticle}
\endbibitem

%%% 3
\bibitem[\protect\citeauthoryear{{Barredo Arrieta}
  et~al.}{2020}]{barredo2020explainable}
\begin{barticle}
\bauthor{\bsnm{{Barredo Arrieta}}, \binits{A.}},
\bauthor{\bsnm{Díaz-Rodríguez}, \binits{N.}},
\bauthor{\bsnm{{Del Ser}}, \binits{J.}},
\bauthor{\bsnm{Bennetot}, \binits{A.}},
\bauthor{\bsnm{Tabik}, \binits{S.}},
\bauthor{\bsnm{Barbado}, \binits{A.}},
\bauthor{\bsnm{Garcia}, \binits{S.}},
\bauthor{\bsnm{Gil-Lopez}, \binits{S.}},
\bauthor{\bsnm{Molina}, \binits{D.}},
\bauthor{\bsnm{Benjamins}, \binits{R.}},
\bauthor{\bsnm{Chatila}, \binits{R.}},
\bauthor{\bsnm{Herrera}, \binits{F.}}:
\batitle{Explainable artificial intelligence (xai): Concepts, taxonomies,
  opportunities and challenges toward responsible ai}.
\bjtitle{Information Fusion}
\bvolume{58},
\bfpage{82}--\blpage{115}
(\byear{2020})
\doiurl{10.1016/j.inffus.2019.12.012}
\end{barticle}
\endbibitem

%%% 4
\bibitem[\protect\citeauthoryear{Montavon et~al.}{2018}]{montavon2018methods}
\begin{barticle}
\bauthor{\bsnm{Montavon}, \binits{G.}},
\bauthor{\bsnm{Samek}, \binits{W.}},
\bauthor{\bsnm{Müller}, \binits{K.-R.}}:
\batitle{Methods for interpreting and understanding deep neural networks}.
\bjtitle{Digital Signal Processing}
\bvolume{73},
\bfpage{1}--\blpage{15}
(\byear{2018})
\doiurl{10.1016/j.dsp.2017.10.011}
\end{barticle}
\endbibitem

%%% 5
\bibitem[\protect\citeauthoryear{Slack et~al.}{2021}]{slack2021advances}
\begin{bchapter}
\bauthor{\bsnm{Slack}, \binits{D.}},
\bauthor{\bsnm{Hilgard}, \binits{A.}},
\bauthor{\bsnm{Singh}, \binits{S.}},
\bauthor{\bsnm{Lakkaraju}, \binits{H.}}:
\bctitle{Reliable post hoc explanations: Modeling uncertainty in
  explainability}.
In: \beditor{\bsnm{Ranzato}, \binits{M.}},
\beditor{\bsnm{Beygelzimer}, \binits{A.}},
\beditor{\bsnm{Dauphin}, \binits{Y.}},
\beditor{\bsnm{Liang}, \binits{P.S.}},
\beditor{\bsnm{Vaughan}, \binits{J.W.}} (eds.)
\bbtitle{Advances in Neural Information Processing Systems},
vol. \bseriesno{34},
pp. \bfpage{9391}--\blpage{9404}.
\bpublisher{Curran Associates, Inc.},
\blocation{Virtual conference}
(\byear{2021}).
\burl{https://proceedings.neurips.cc/paper_files/paper/2021/file/4e246a381baf2ce038b3b0f82c7d6fb4-Paper.pdf}
\end{bchapter}
\endbibitem

%%% 6
\bibitem[\protect\citeauthoryear{Kotsiantis}{2013}]{kotsiantis2013decision}
\begin{barticle}
\bauthor{\bsnm{Kotsiantis}, \binits{S.B.}}:
\batitle{Decision trees: a recent overview}.
\bjtitle{Artificial Intelligence Review}
\bvolume{39}(\bissue{4}),
\bfpage{261}--\blpage{283}
(\byear{2013})
\doiurl{10.1007/s10462-011-9272-4}
\end{barticle}
\endbibitem

%%% 7
\bibitem[\protect\citeauthoryear{Breiman}{2001}]{breiman2001random}
\begin{barticle}
\bauthor{\bsnm{Breiman}, \binits{L.}}:
\batitle{Random forests}.
\bjtitle{Machine learning}
\bvolume{45},
\bfpage{5}--\blpage{32}
(\byear{2001})
\doiurl{10.1023/A:1010933404324}
\end{barticle}
\endbibitem

%%% 8
\bibitem[\protect\citeauthoryear{Chen and Guestrin}{2016}]{chen2016xgboost}
\begin{bchapter}
\bauthor{\bsnm{Chen}, \binits{T.}},
\bauthor{\bsnm{Guestrin}, \binits{C.}}:
\bctitle{Xgboost: A scalable tree boosting system}.
In: \bbtitle{Proceedings of the 22nd ACM SIGKDD International Conference on
  Knowledge Discovery and Data Mining}.
\bsertitle{KDD '16},
pp. \bfpage{785}--\blpage{794}.
\bpublisher{Association for Computing Machinery},
\blocation{New York, NY, USA}
(\byear{2016}).
\doiurl{10.1145/2939672.2939785} .
\burl{https://doi.org/10.1145/2939672.2939785}
\end{bchapter}
\endbibitem

%%% 9
\bibitem[\protect\citeauthoryear{Mosqueira-Rey
  et~al.}{2023}]{mosqueira2023human}
\begin{barticle}
\bauthor{\bsnm{Mosqueira-Rey}, \binits{E.}},
\bauthor{\bsnm{Hern\'andez-Pereira}, \binits{E.}},
\bauthor{\bsnm{Alonso-R\'ios}, \binits{D.}},
\bauthor{\bsnm{Bobes-Bascar\'an}, \binits{J.}},
\bauthor{\bsnm{Fern\'andez-Leal}, \binits{A.}}:
\batitle{Human-in-the-loop machine learning: A state of the art}.
\bjtitle{Artificial Intelligence Review}
\bvolume{3005},
\bfpage{3054}
(\byear{2023})
\doiurl{10.1007/s10462-022-10246-w}
\end{barticle}
\endbibitem

%%% 10
\bibitem[\protect\citeauthoryear{Gilpin et~al.}{2018}]{gilpin2018explaining}
\begin{bchapter}
\bauthor{\bsnm{Gilpin}, \binits{L.H.}},
\bauthor{\bsnm{Bau}, \binits{D.}},
\bauthor{\bsnm{Yuan}, \binits{B.Z.}},
\bauthor{\bsnm{Bajwa}, \binits{A.}},
\bauthor{\bsnm{Specter}, \binits{M.}},
\bauthor{\bsnm{Kagal}, \binits{L.}}:
\bctitle{Explaining explanations: An overview of interpretability of machine
  learning}.
In: \bbtitle{2018 IEEE 5th International Conference on Data Science and
  Advanced Analytics (DSAA)},
pp. \bfpage{80}--\blpage{89}
(\byear{2018}).
\doiurl{10.1109/DSAA.2018.00018}
\end{bchapter}
\endbibitem

%%% 11
\bibitem[\protect\citeauthoryear{Lipton}{2018}]{lipton2018mythos}
\begin{barticle}
\bauthor{\bsnm{Lipton}, \binits{Z.C.}}:
\batitle{The mythos of model interpretability: In machine learning, the concept
  of interpretability is both important and slippery.}
\bjtitle{Queue}
\bvolume{16}(\bissue{3}),
\bfpage{31}--\blpage{57}
(\byear{2018})
\doiurl{10.48550/arXiv.1606.03490}
\end{barticle}
\endbibitem

%%% 12
\bibitem[\protect\citeauthoryear{Carrillo
  et~al.}{2021}]{carrillo2021individual}
\begin{botherref}
\oauthor{\bsnm{Carrillo}, \binits{A.}},
\oauthor{\bsnm{Cantú}, \binits{L.F.}},
\oauthor{\bsnm{Noriega}, \binits{A.}}:
Individual Explanations in Machine Learning Models: A Survey for Practitioners
(2021).
\doiurl{10.48550/arXiv.2104.04144}
\end{botherref}
\endbibitem

%%% 13
\bibitem[\protect\citeauthoryear{Goodman and
  Flaxman}{2017}]{goodman2017european}
\begin{barticle}
\bauthor{\bsnm{Goodman}, \binits{B.}},
\bauthor{\bsnm{Flaxman}, \binits{S.}}:
\batitle{European union regulations on algorithmic decision-making and a
  “right to explanation”}.
\bjtitle{AI magazine}
\bvolume{38}(\bissue{3}),
\bfpage{50}--\blpage{57}
(\byear{2017})
\end{barticle}
\endbibitem

%%% 14
\bibitem[\protect\citeauthoryear{Adadi and Berrada}{2018}]{adadi2018peeking}
\begin{barticle}
\bauthor{\bsnm{Adadi}, \binits{A.}},
\bauthor{\bsnm{Berrada}, \binits{M.}}:
\batitle{Peeking inside the black-box: A survey on explainable artificial
  intelligence (xai)}.
\bjtitle{IEEE Access}
\bvolume{6},
\bfpage{52138}--\blpage{52160}
(\byear{2018})
\doiurl{10.1109/ACCESS.2018.2870052}
\end{barticle}
\endbibitem

%%% 15
\bibitem[\protect\citeauthoryear{Holzinger
  et~al.}{2019}]{holzinger2019causability}
\begin{barticle}
\bauthor{\bsnm{Holzinger}, \binits{A.}},
\bauthor{\bsnm{Langs}, \binits{G.}},
\bauthor{\bsnm{Denk}, \binits{H.}},
\bauthor{\bsnm{Zatloukal}, \binits{K.}},
\bauthor{\bsnm{Müller}, \binits{H.}}:
\batitle{Causability and explainability of artificial intelligence in
  medicine}.
\bjtitle{WIREs Data Mining and Knowledge Discovery}
\bvolume{9}(\bissue{4}),
\bfpage{1312}
(\byear{2019})
\doiurl{10.1002/widm.1312}
{\href{https://arxiv.org/abs/https://wires.onlinelibrary.wiley.com/doi/pdf/10.1002/widm.1312}{{https://wires.onlinelibrary.wiley.com/doi/pdf/10.1002/widm.1312}}}
\end{barticle}
\endbibitem

%%% 16
\bibitem[\protect\citeauthoryear{Holzinger et~al.}{2017}]{holzinger2017what}
\begin{botherref}
\oauthor{\bsnm{Holzinger}, \binits{A.}},
\oauthor{\bsnm{Biemann}, \binits{C.}},
\oauthor{\bsnm{Pattichis}, \binits{C.S.}},
\oauthor{\bsnm{Kell}, \binits{D.B.}}:
What do we need to build explainable {AI} systems for the medical domain?
arXiv preprint arXiv:1712.09923
(2017)
\end{botherref}
\endbibitem

%%% 17
\bibitem[\protect\citeauthoryear{Loh et~al.}{2022}]{loh2022application}
\begin{barticle}
\bauthor{\bsnm{Loh}, \binits{H.W.}},
\bauthor{\bsnm{Ooi}, \binits{C.P.}},
\bauthor{\bsnm{Seoni}, \binits{S.}},
\bauthor{\bsnm{Barua}, \binits{P.D.}},
\bauthor{\bsnm{Molinari}, \binits{F.}},
\bauthor{\bsnm{Acharya}, \binits{U.R.}}:
\batitle{Application of explainable artificial intelligence for healthcare: A
  systematic review of the last decade (2011–2022)}.
\bjtitle{Computer Methods and Programs in Biomedicine}
\bvolume{226},
\bfpage{107161}
(\byear{2022})
\doiurl{10.1016/j.cmpb.2022.107161}
\end{barticle}
\endbibitem

%%% 18
\bibitem[\protect\citeauthoryear{Nair et~al.}{2023}]{nair2023building}
\begin{barticle}
\bauthor{\bsnm{Nair}, \binits{P.C.}},
\bauthor{\bsnm{Gupta}, \binits{D.}},
\bauthor{\bsnm{Devi}, \binits{B.I.}},
\bauthor{\bsnm{Kanjirangat}, \binits{V.}}:
\batitle{Building an explainable diagnostic classification model for brain
  tumor using discharge summaries}.
\bjtitle{Procedia Computer Science}
\bvolume{218},
\bfpage{2058}--\blpage{2070}
(\byear{2023})
\end{barticle}
\endbibitem

%%% 19
\bibitem[\protect\citeauthoryear{Ganeshkumar
  et~al.}{2021}]{ganeshkumar2021explainable}
\begin{barticle}
\bauthor{\bsnm{Ganeshkumar}, \binits{M.}},
\bauthor{\bsnm{Ravi}, \binits{V.}},
\bauthor{\bsnm{Sowmya}, \binits{V.}},
\bauthor{\bsnm{Gopalakrishnan}, \binits{E.}},
\bauthor{\bsnm{Soman}, \binits{K.}}:
\batitle{Explainable deep learning-based approach for multilabel classification
  of electrocardiogram}.
\bjtitle{IEEE Transactions on Engineering Management}
(\byear{2021})
\doiurl{10.1109/TEM.2021.3104751}
\end{barticle}
\endbibitem

%%% 20
\bibitem[\protect\citeauthoryear{Moncada-Torres
  et~al.}{2021}]{moncada2021explainable}
\begin{barticle}
\bauthor{\bsnm{Moncada-Torres}, \binits{A.}},
\bauthor{\bsnm{Maaren}, \binits{M.C.}},
\bauthor{\bsnm{Hendriks}, \binits{M.P.}},
\bauthor{\bsnm{Siesling}, \binits{S.}},
\bauthor{\bsnm{Geleijnse}, \binits{G.}}:
\batitle{Explainable machine learning can outperform cox regression predictions
  and provide insights in breast cancer survival}.
\bjtitle{Scientific reports}
\bvolume{11}(\bissue{1}),
\bfpage{6968}
(\byear{2021})
\doiurl{10.1038/s41598-021-86327-7}
\end{barticle}
\endbibitem

%%% 21
\bibitem[\protect\citeauthoryear{Gulum et~al.}{2021}]{gulum2021review}
\begin{barticle}
\bauthor{\bsnm{Gulum}, \binits{M.A.}},
\bauthor{\bsnm{Trombley}, \binits{C.M.}},
\bauthor{\bsnm{Kantardzic}, \binits{M.}}:
\batitle{A review of explainable deep learning cancer detection models in
  medical imaging}.
\bjtitle{Applied Sciences}
\bvolume{11}(\bissue{10}),
\bfpage{4573}
(\byear{2021})
\end{barticle}
\endbibitem

%%% 22
\bibitem[\protect\citeauthoryear{Hauser et~al.}{2022}]{hauser2022explainable}
\begin{barticle}
\bauthor{\bsnm{Hauser}, \binits{K.}},
\bauthor{\bsnm{Kurz}, \binits{A.}},
\bauthor{\bsnm{Haggenmueller}, \binits{S.}},
\bauthor{\bsnm{Maron}, \binits{R.C.}},
\bauthor{\bsnm{Kalle}, \binits{C.}},
\bauthor{\bsnm{Utikal}, \binits{J.S.}},
\bauthor{\bsnm{Meier}, \binits{F.}},
\bauthor{\bsnm{Hobelsberger}, \binits{S.}},
\bauthor{\bsnm{Gellrich}, \binits{F.F.}},
\bauthor{\bsnm{Sergon}, \binits{M.}}, \betal:
\batitle{Explainable artificial intelligence in skin cancer recognition: A
  systematic review}.
\bjtitle{European Journal of Cancer}
\bvolume{167},
\bfpage{54}--\blpage{69}
(\byear{2022})
\end{barticle}
\endbibitem

%%% 23
\bibitem[\protect\citeauthoryear{Bhatt et~al.}{2020}]{bhatt2020explainable}
\begin{bchapter}
\bauthor{\bsnm{Bhatt}, \binits{U.}},
\bauthor{\bsnm{Xiang}, \binits{A.}},
\bauthor{\bsnm{Sharma}, \binits{S.}},
\bauthor{\bsnm{Weller}, \binits{A.}},
\bauthor{\bsnm{Taly}, \binits{A.}},
\bauthor{\bsnm{Jia}, \binits{Y.}},
\bauthor{\bsnm{Ghosh}, \binits{J.}},
\bauthor{\bsnm{Puri}, \binits{R.}},
\bauthor{\bsnm{Moura}, \binits{J.M.}},
\bauthor{\bsnm{Eckersley}, \binits{P.}}:
\bctitle{Explainable machine learning in deployment}.
In: \bbtitle{Proceedings of the 2020 Conference on Fairness, Accountability,
  and Transparency},
pp. \bfpage{648}--\blpage{657}
(\byear{2020})
\end{bchapter}
\endbibitem

%%% 24
\bibitem[\protect\citeauthoryear{Dimanov et~al.}{2020}]{dimanov2020you}
\begin{botherref}
\oauthor{\bsnm{Dimanov}, \binits{B.}},
\oauthor{\bsnm{Bhatt}, \binits{U.}},
\oauthor{\bsnm{Jamnik}, \binits{M.}},
\oauthor{\bsnm{Weller}, \binits{A.}}:
You shouldn’t trust me: Learning models which conceal unfairness from
  multiple explanation methods.
(2020)
\end{botherref}
\endbibitem

%%% 25
\bibitem[\protect\citeauthoryear{Ghorbani
  et~al.}{2019}]{ghorbani2019interpretation}
\begin{bchapter}
\bauthor{\bsnm{Ghorbani}, \binits{A.}},
\bauthor{\bsnm{Abid}, \binits{A.}},
\bauthor{\bsnm{Zou}, \binits{J.}}:
\bctitle{Interpretation of neural networks is fragile}.
In: \bbtitle{Proceedings of the AAAI Conference on Artificial Intelligence},
vol. \bseriesno{33},
pp. \bfpage{3681}--\blpage{3688}
(\byear{2019}).
\doiurl{10.1609/aaai.v33i01.33013681} .
\burl{https://ojs.aaai.org/index.php/AAAI/article/view/4252}
\end{bchapter}
\endbibitem

%%% 26
\bibitem[\protect\citeauthoryear{Slack et~al.}{2020}]{slack2020fooling}
\begin{bchapter}
\bauthor{\bsnm{Slack}, \binits{D.}},
\bauthor{\bsnm{Hilgard}, \binits{S.}},
\bauthor{\bsnm{Jia}, \binits{E.}},
\bauthor{\bsnm{Singh}, \binits{S.}},
\bauthor{\bsnm{Lakkaraju}, \binits{H.}}:
\bctitle{Fooling lime and shap: Adversarial attacks on post hoc explanation
  methods}.
In: \bbtitle{Proceedings of the AAAI/ACM Conference on AI, Ethics, and
  Society}.
\bsertitle{AIES '20},
pp. \bfpage{180}--\blpage{186}.
\bpublisher{Association for Computing Machinery},
\blocation{New York, NY, USA}
(\byear{2020}).
\doiurl{10.1145/3375627.3375830}
\end{bchapter}
\endbibitem

%%% 27
\bibitem[\protect\citeauthoryear{Dombrowski
  et~al.}{2019}]{dombrowski2019explanations}
\begin{botherref}
\oauthor{\bsnm{Dombrowski}, \binits{A.}},
\oauthor{\bsnm{Alber}, \binits{M.}},
\oauthor{\bsnm{Anders}, \binits{C.J.}},
\oauthor{\bsnm{Ackermann}, \binits{M.}},
\oauthor{\bsnm{M{\"{u}}ller}, \binits{K.}},
\oauthor{\bsnm{Kessel}, \binits{P.}}:
Explanations can be manipulated and geometry is to blame.
CoRR
\textbf{abs/1906.07983}
(2019)
{\href{https://arxiv.org/abs/1906.07983}{{1906.07983}}}
\end{botherref}
\endbibitem

%%% 28
\bibitem[\protect\citeauthoryear{Alvarez{-}Melis and
  Jaakkola}{2018}]{alvarez2018robustness}
\begin{botherref}
\oauthor{\bsnm{Alvarez{-}Melis}, \binits{D.}},
\oauthor{\bsnm{Jaakkola}, \binits{T.S.}}:
On the robustness of interpretability methods.
CoRR
\textbf{abs/1806.08049}
(2018)
{\href{https://arxiv.org/abs/1806.08049}{{1806.08049}}}
\end{botherref}
\endbibitem

%%% 29
\bibitem[\protect\citeauthoryear{Lee et~al.}{2019}]{lee2019developing}
\begin{bchapter}
\bauthor{\bsnm{Lee}, \binits{E.}},
\bauthor{\bsnm{Braines}, \binits{D.}},
\bauthor{\bsnm{Stiffler}, \binits{M.}},
\bauthor{\bsnm{Hudler}, \binits{A.}},
\bauthor{\bsnm{Harborne}, \binits{D.}}:
\bctitle{{Developing the sensitivity of LIME for better machine learning
  explanation}}.
In: \beditor{\bsnm{Pham}, \binits{T.}} (ed.)
\bbtitle{Artificial Intelligence and Machine Learning for Multi-Domain
  Operations Applications},
vol. \bseriesno{11006},
pp. \bfpage{349}--\blpage{356}.
\bpublisher{SPIE},
\blocation{Baltimore, MD, USA}
(\byear{2019}).
\doiurl{10.1117/12.2520149} .
\bcomment{International Society for Optics and Photonics}.
\burl{https://doi.org/10.1117/12.2520149}
\end{bchapter}
\endbibitem

%%% 30
\bibitem[\protect\citeauthoryear{Tan et~al.}{2019}]{fen2019why}
\begin{botherref}
\oauthor{\bsnm{Tan}, \binits{H.F.}},
\oauthor{\bsnm{Song}, \binits{K.}},
\oauthor{\bsnm{Udell}, \binits{M.}},
\oauthor{\bsnm{Sun}, \binits{Y.}},
\oauthor{\bsnm{Zhang}, \binits{Y.}}:
Why should you trust my interpretation? understanding uncertainty in {LIME}
  predictions.
CoRR
\textbf{abs/1904.12991}
(2019)
{\href{https://arxiv.org/abs/1904.12991}{{1904.12991}}}
\end{botherref}
\endbibitem

%%% 31
\bibitem[\protect\citeauthoryear{Zafar and Khan}{2019}]{zafar2019dlime}
\begin{botherref}
\oauthor{\bsnm{Zafar}, \binits{M.R.}},
\oauthor{\bsnm{Khan}, \binits{N.M.}}:
{DLIME:} {A} deterministic local interpretable model-agnostic explanations
  approach for computer-aided diagnosis systems.
CoRR
\textbf{abs/1906.10263}
(2019)
{\href{https://arxiv.org/abs/1906.10263}{{1906.10263}}}
\end{botherref}
\endbibitem

%%% 32
\bibitem[\protect\citeauthoryear{Bray et~al.}{2018}]{bray2018global}
\begin{barticle}
\bauthor{\bsnm{Bray}, \binits{F.}},
\bauthor{\bsnm{Ferlay}, \binits{J.}},
\bauthor{\bsnm{Soerjomataram}, \binits{I.}},
\bauthor{\bsnm{Siegel}, \binits{R.L.}},
\bauthor{\bsnm{Torre}, \binits{L.A.}},
\bauthor{\bsnm{Jemal}, \binits{A.}}:
\batitle{Global cancer statistics 2018: Globocan estimates of incidence and
  mortality worldwide for 36 cancers in 185 countries}.
\bjtitle{CA: a cancer journal for clinicians}
\bvolume{68}(\bissue{6}),
\bfpage{394}--\blpage{424}
(\byear{2018})
\end{barticle}
\endbibitem

%%% 33
\bibitem[\protect\citeauthoryear{Tomczak et~al.}{2015}]{tomczak2015cancer}
\begin{barticle}
\bauthor{\bsnm{Tomczak}, \binits{K.}},
\bauthor{\bsnm{Czerwi{\'n}ska}, \binits{P.}},
\bauthor{\bsnm{Wiznerowicz}, \binits{M.}}:
\batitle{The cancer genome atlas ({TCGA}): an immeasurable source of
  knowledge}.
\bjtitle{Contemporary Oncology}
\bvolume{19}(\bissue{1A}),
\bfpage{68}--\blpage{77}
(\byear{2015})
\doiurl{10.5114/wo.2014.47136}
\end{barticle}
\endbibitem

%%% 34
\bibitem[\protect\citeauthoryear{McGuigan
  et~al.}{2018}]{mcguigan2018pancreatic}
\begin{barticle}
\bauthor{\bsnm{McGuigan}, \binits{A.}},
\bauthor{\bsnm{Kelly}, \binits{P.}},
\bauthor{\bsnm{Turkington}, \binits{R.C.}},
\bauthor{\bsnm{Jones}, \binits{C.}},
\bauthor{\bsnm{Coleman}, \binits{H.G.}},
\bauthor{\bsnm{McCain}, \binits{R.S.}}:
\batitle{Pancreatic cancer: A review of clinical diagnosis, epidemiology,
  treatment and outcomes}.
\bjtitle{World journal of gastroenterology}
\bvolume{24}(\bissue{43}),
\bfpage{4846}
(\byear{2018})
\end{barticle}
\endbibitem

%%% 35
\bibitem[\protect\citeauthoryear{Hunter et~al.}{2022}]{hunter2022role}
\begin{botherref}
\oauthor{\bsnm{Hunter}, \binits{B.}},
\oauthor{\bsnm{Hindocha}, \binits{S.}},
\oauthor{\bsnm{Lee}, \binits{R.W.}}:
The role of artificial intelligence in early cancer diagnosis.
Cancers
\textbf{14}(6)
(2022)
\doiurl{10.3390/cancers14061524}
\end{botherref}
\endbibitem

%%% 36
\bibitem[\protect\citeauthoryear{Kenner et~al.}{2021}]{kenner2021artificial}
\begin{barticle}
\bauthor{\bsnm{Kenner}, \binits{B.}},
\bauthor{\bsnm{Chari}, \binits{S.T.}},
\bauthor{\bsnm{Kelsen}, \binits{D.}},
\bauthor{\bsnm{Klimstra}, \binits{D.S.}},
\bauthor{\bsnm{Pandol}, \binits{S.J.}},
\bauthor{\bsnm{Rosenthal}, \binits{M.}},
\bauthor{\bsnm{Rustgi}, \binits{A.K.}},
\bauthor{\bsnm{Taylor}, \binits{J.A.}},
\bauthor{\bsnm{Yala}, \binits{A.}},
\bauthor{\bsnm{Abul-Husn}, \binits{N.}}, \betal:
\batitle{Artificial intelligence and early detection of pancreatic cancer: 2020
  summative review}.
\bjtitle{Pancreas}
\bvolume{50}(\bissue{3}),
\bfpage{251}
(\byear{2021})
\end{barticle}
\endbibitem

%%% 37
\bibitem[\protect\citeauthoryear{Dmitriev
  et~al.}{2021}]{dmitriev2021visualAnalytics}
\begin{barticle}
\bauthor{\bsnm{Dmitriev}, \binits{K.}},
\bauthor{\bsnm{Marino}, \binits{J.}},
\bauthor{\bsnm{Baker}, \binits{K.}},
\bauthor{\bsnm{Kaufman}, \binits{A.E.}}:
\batitle{Visual analytics of a computer-aided diagnosis system for pancreatic
  lesions}.
\bjtitle{IEEE Transactions on Visualization and Computer Graphics}
\bvolume{27}(\bissue{3}),
\bfpage{2174}--\blpage{2185}
(\byear{2021})
\doiurl{10.1109/TVCG.2019.2947037}
\end{barticle}
\endbibitem

%%% 38
\bibitem[\protect\citeauthoryear{Bakasa and
  Viriri}{2021}]{bakasa2021survivalPrediction}
\begin{barticle}
\bauthor{\bsnm{Bakasa}, \binits{W.}},
\bauthor{\bsnm{Viriri}, \binits{S.}}:
\batitle{Pancreatic cancer survival prediction: a survey of the
  state-of-the-art}.
\bjtitle{Computational and Mathematical Methods in Medicine}
\bvolume{2021},
\bfpage{1}--\blpage{17}
(\byear{2021})
\end{barticle}
\endbibitem

%%% 39
\bibitem[\protect\citeauthoryear{Walczak and
  Velanovich}{2017}]{walczak2017evaluation}
\begin{barticle}
\bauthor{\bsnm{Walczak}, \binits{S.}},
\bauthor{\bsnm{Velanovich}, \binits{V.}}:
\batitle{An evaluation of artificial neural networks in predicting pancreatic
  cancer survival}.
\bjtitle{Journal of Gastrointestinal Surgery}
\bvolume{21},
\bfpage{1606}--\blpage{1612}
(\byear{2017})
\doiurl{10.1007/s11605-017-3518-7}
\end{barticle}
\endbibitem

%%% 40
\bibitem[\protect\citeauthoryear{Hayashi et~al.}{2021}]{hayashi2021recent}
\begin{barticle}
\bauthor{\bsnm{Hayashi}, \binits{H.}},
\bauthor{\bsnm{Uemura}, \binits{N.}},
\bauthor{\bsnm{Matsumura}, \binits{K.}},
\bauthor{\bsnm{Zhao}, \binits{L.}},
\bauthor{\bsnm{Sato}, \binits{H.}},
\bauthor{\bsnm{Shiraishi}, \binits{Y.}},
\bauthor{\bsnm{Yamashita}, \binits{Y.-i.}},
\bauthor{\bsnm{Baba}, \binits{H.}}:
\batitle{Recent advances in artificial intelligence for pancreatic ductal
  adenocarcinoma}.
\bjtitle{World Journal of Gastroenterology}
\bvolume{27}(\bissue{43}),
\bfpage{7480}
(\byear{2021})
\end{barticle}
\endbibitem

%%% 41
\bibitem[\protect\citeauthoryear{Bradley
  et~al.}{2019}]{bradley2019personalized}
\begin{barticle}
\bauthor{\bsnm{Bradley}, \binits{A.}},
\bauthor{\bsnm{Van Der~Meer}, \binits{R.}},
\bauthor{\bsnm{McKay}, \binits{C.}}:
\batitle{Personalized pancreatic cancer management: a systematic review of how
  machine learning is supporting decision-making}.
\bjtitle{Pancreas}
\bvolume{48}(\bissue{5}),
\bfpage{598}--\blpage{604}
(\byear{2019})
\end{barticle}
\endbibitem

%%% 42
\bibitem[\protect\citeauthoryear{Amin et~al.}{2017}]{amin2017eighth}
\begin{barticle}
\bauthor{\bsnm{Amin}, \binits{M.B.}},
\bauthor{\bsnm{Greene}, \binits{F.L.}},
\bauthor{\bsnm{Edge}, \binits{S.B.}},
\bauthor{\bsnm{Compton}, \binits{C.C.}},
\bauthor{\bsnm{Gershenwald}, \binits{J.E.}},
\bauthor{\bsnm{Brookland}, \binits{R.K.}},
\bauthor{\bsnm{Meyer}, \binits{L.}},
\bauthor{\bsnm{Gress}, \binits{D.M.}},
\bauthor{\bsnm{Byrd}, \binits{D.R.}},
\bauthor{\bsnm{Winchester}, \binits{D.P.}}:
\batitle{The eighth edition ajcc cancer staging manual: continuing to build a
  bridge from a population-based to a more “personalized” approach to
  cancer staging}.
\bjtitle{CA: a cancer journal for clinicians}
\bvolume{67}(\bissue{2}),
\bfpage{93}--\blpage{99}
(\byear{2017})
\end{barticle}
\endbibitem

%%% 43
\bibitem[\protect\citeauthoryear{Cong et~al.}{2018}]{cong2018tumor}
\begin{barticle}
\bauthor{\bsnm{Cong}, \binits{L.}},
\bauthor{\bsnm{Liu}, \binits{Q.}},
\bauthor{\bsnm{Zhang}, \binits{R.}},
\bauthor{\bsnm{Cui}, \binits{M.}},
\bauthor{\bsnm{Zhang}, \binits{X.}},
\bauthor{\bsnm{Gao}, \binits{X.}},
\bauthor{\bsnm{Guo}, \binits{J.}},
\bauthor{\bsnm{Dai}, \binits{M.}},
\bauthor{\bsnm{Zhang}, \binits{T.}},
\bauthor{\bsnm{Liao}, \binits{Q.}}, \betal:
\batitle{Tumor size classification of the 8th edition of tnm staging system is
  superior to that of the 7th edition in predicting the survival outcome of
  pancreatic cancer patiments after radical resection and adjuvant
  chemotherapy}.
\bjtitle{Scientific reports}
\bvolume{8}(\bissue{1}),
\bfpage{10383}
(\byear{2018})
\end{barticle}
\endbibitem

%%% 44
\bibitem[\protect\citeauthoryear{NCCN}{2022}]{nccn2022pancreatic}
\begin{bbook}
\bauthor{\bsnm{NCCN}}:
\bbtitle{Pancreatic Adenocarcinoma, Version 3.2019}.
\bpublisher{National Comprehensive Cancer Network},
\blocation{Plymouth Meeting, PA}
(\byear{2022})
\end{bbook}
\endbibitem

%%% 45
\bibitem[\protect\citeauthoryear{{Cancer Genome Atlas Research Network}
  et~al.}{2013}]{CancerGenomeAtlasResearchNetwork:2013}
\begin{barticle}
\bauthor{\bsnm{{Cancer Genome Atlas Research Network}}},
\bauthor{\bsnm{Weinstein}, \binits{J.N.}},
\bauthor{\bsnm{Collisson}, \binits{E.A.}},
\bauthor{\bsnm{Mills}, \binits{G.B.}},
\bauthor{\bsnm{Shaw}, \binits{K.R.}},
\bauthor{\bsnm{Ozenberger}, \binits{B.A.}},
\bauthor{\bsnm{Ellrott}, \binits{K.}},
\bauthor{\bsnm{Shmulevich}, \binits{I.}},
\bauthor{\bsnm{Sander}, \binits{C.}},
\bauthor{\bsnm{Stuart}, \binits{J.M.}}:
\batitle{The cancer genome atlas pan-cancer analysis project}.
\bjtitle{Nat Genet}
\bvolume{45}(\bissue{10}),
\bfpage{1113}--\blpage{1120}
(\byear{2013})
\doiurl{10.1038/ng.2764}
\end{barticle}
\endbibitem

%%% 46
\bibitem[\protect\citeauthoryear{Mosqueira-Rey
  et~al.}{2024}]{mosqueira2024addressing}
\begin{barticle}
\bauthor{\bsnm{Mosqueira-Rey}, \binits{E.}},
\bauthor{\bsnm{Hern\'andez-Pereira}, \binits{E.}},
\bauthor{\bsnm{Bobes-Bascar\'an}, \binits{J.}},
\bauthor{\bsnm{Alonso-R\'ios}, \binits{D.}},
\bauthor{\bsnm{P\'erez-S\'anchez}, \binits{A.}},
\bauthor{\bsnm{Fern\'andez-Leal}, \binits{A.}},
\bauthor{\bsnm{Moret-Bonillo}, \binits{V.}},
\bauthor{\bsnm{Vidal-\'Insua}, \binits{Y.}},
\bauthor{\bsnm{V\'azquez-Rivera}, \binits{F.}}:
\batitle{Addressing the data bottleneck in medical deep learning models using a
  human-in-the-loop machine learning approach}.
\bjtitle{Neural Computing and Applications}
\bvolume{36}(\bissue{5}),
\bfpage{2597}--\blpage{2616}
(\byear{2024})
\doiurl{10.1007/s00521-023-09197-2}
\end{barticle}
\endbibitem

%%% 47
\bibitem[\protect\citeauthoryear{Samaan et~al.}{2023}]{samaan2023pancreatic}
\begin{barticle}
\bauthor{\bsnm{Samaan}, \binits{J.S.}},
\bauthor{\bsnm{Abboud}, \binits{Y.}},
\bauthor{\bsnm{Oh}, \binits{J.}},
\bauthor{\bsnm{Jiang}, \binits{Y.}},
\bauthor{\bsnm{Watson}, \binits{R.}},
\bauthor{\bsnm{Park}, \binits{K.}},
\bauthor{\bsnm{Liu}, \binits{Q.}},
\bauthor{\bsnm{Atkins}, \binits{K.}},
\bauthor{\bsnm{Hendifar}, \binits{A.}},
\bauthor{\bsnm{Gong}, \binits{J.}}, \betal:
\batitle{Pancreatic cancer incidence trends by race, ethnicity, age and sex in
  the united states: a population-based study, 2000--2018}.
\bjtitle{Cancers}
\bvolume{15}(\bissue{3}),
\bfpage{870}
(\byear{2023})
\end{barticle}
\endbibitem

%%% 48
\bibitem[\protect\citeauthoryear{Tempero
  et~al.}{2021}]{nccnClinicalPractice2021}
\begin{barticle}
\bauthor{\bsnm{Tempero}, \binits{M.A.}},
\bauthor{\bsnm{Malafa}, \binits{M.P.}},
\bauthor{\bsnm{Al-Hawary}, \binits{M.}},
\bauthor{\bsnm{Behrman}, \binits{S.W.}},
\bauthor{\bsnm{Benson}, \binits{A.B.}},
\bauthor{\bsnm{Cardin}, \binits{D.B.}},
\bauthor{\bsnm{Chiorean}, \binits{E.G.}},
\bauthor{\bsnm{Chung}, \binits{V.}},
\bauthor{\bsnm{Czito}, \binits{B.}},
\bauthor{\bsnm{Chiaro}, \binits{M.D.}},
\bauthor{\bsnm{Dillhoff}, \binits{M.}},
\bauthor{\bsnm{Donahue}, \binits{T.R.}},
\bauthor{\bsnm{Dotan}, \binits{E.}},
\bauthor{\bsnm{Ferrone}, \binits{C.R.}},
\bauthor{\bsnm{Fountzilas}, \binits{C.}},
\bauthor{\bsnm{Hardacre}, \binits{J.}},
\bauthor{\bsnm{Hawkins}, \binits{W.G.}},
\bauthor{\bsnm{Klute}, \binits{K.}},
\bauthor{\bsnm{Ko}, \binits{A.H.}},
\bauthor{\bsnm{Kunstman}, \binits{J.W.}},
\bauthor{\bsnm{LoConte}, \binits{N.}},
\bauthor{\bsnm{Lowy}, \binits{A.M.}},
\bauthor{\bsnm{Moravek}, \binits{C.}},
\bauthor{\bsnm{Nakakura}, \binits{E.K.}},
\bauthor{\bsnm{Narang}, \binits{A.K.}},
\bauthor{\bsnm{Obando}, \binits{J.}},
\bauthor{\bsnm{Polanco}, \binits{P.M.}},
\bauthor{\bsnm{Reddy}, \binits{S.}},
\bauthor{\bsnm{Reyngold}, \binits{M.}},
\bauthor{\bsnm{Scaife}, \binits{C.}},
\bauthor{\bsnm{Shen}, \binits{J.}},
\bauthor{\bsnm{Vollmer}, \binits{C.}},
\bauthor{\bsnm{Wolff}, \binits{R.A.}},
\bauthor{\bsnm{Wolpin}, \binits{B.M.}},
\bauthor{\bsnm{Lynn}, \binits{B.}},
\bauthor{\bsnm{George}, \binits{G.V.}}:
\batitle{Pancreatic adenocarcinoma, version 2.2021, nccn clinical practice
  guidelines in oncology}.
\bjtitle{Journal of the National Comprehensive Cancer Network}
\bvolume{19}(\bissue{4}),
\bfpage{439}--\blpage{457}
(\byear{2021})
\doiurl{10.6004/jnccn.2021.0017}
\end{barticle}
\endbibitem

%%% 49
\bibitem[\protect\citeauthoryear{Breiman
  et~al.}{2017}]{breiman2017classification}
\begin{bbook}
\bauthor{\bsnm{Breiman}, \binits{L.}},
\bauthor{\bsnm{Friedman}, \binits{J.}},
\bauthor{\bsnm{Olshen}, \binits{R.A.}},
\bauthor{\bsnm{Stone}, \binits{C.J.}}:
\bbtitle{Classification and Regression Trees}.
\bpublisher{Taylor and Francis Group},
\blocation{New York}
(\byear{2017}).
\doiurl{10.1201/9781315139470}
\end{bbook}
\endbibitem

%%% 50
\bibitem[\protect\citeauthoryear{Bent{\'e}jac
  et~al.}{2021}]{bentejac2021comparative}
\begin{barticle}
\bauthor{\bsnm{Bent{\'e}jac}, \binits{C.}},
\bauthor{\bsnm{Cs{\"o}rg{\H{o}}}, \binits{A.}},
\bauthor{\bsnm{Mart{\'\i}nez-Mu{\~n}oz}, \binits{G.}}:
\batitle{A comparative analysis of gradient boosting algorithms}.
\bjtitle{Artificial Intelligence Review}
\bvolume{54},
\bfpage{1937}--\blpage{1967}
(\byear{2021})
\doiurl{10.1007/s10462-020-09896-5}
\end{barticle}
\endbibitem

%%% 51
\bibitem[\protect\citeauthoryear{Saarela and
  Jauhiainen}{2021}]{saarela2021comparison}
\begin{barticle}
\bauthor{\bsnm{Saarela}, \binits{M.}},
\bauthor{\bsnm{Jauhiainen}, \binits{S.}}:
\batitle{Comparison of feature importance measures as explanations for
  classification models}.
\bjtitle{SN Applied Sciences}
\bvolume{3}(\bissue{2}),
\bfpage{272}
(\byear{2021})
\doiurl{10.1007/s42452-021-04148-9}
\end{barticle}
\endbibitem

%%% 52
\bibitem[\protect\citeauthoryear{Zhou and Hooker}{2021}]{zhou2021unbiased}
\begin{botherref}
\oauthor{\bsnm{Zhou}, \binits{Z.}},
\oauthor{\bsnm{Hooker}, \binits{G.}}:
Unbiased measurement of feature importance in tree-based methods.
ACM Trans. Knowl. Discov. Data
\textbf{15}(2)
(2021)
\doiurl{10.1145/3429445}
\end{botherref}
\endbibitem

%%% 53
\bibitem[\protect\citeauthoryear{Lundberg and Lee}{2017}]{lundberg2017unified}
\begin{bchapter}
\bauthor{\bsnm{Lundberg}, \binits{S.M.}},
\bauthor{\bsnm{Lee}, \binits{S.-I.}}:
\bctitle{A unified approach to interpreting model predictions}.
In: \beditor{\bsnm{Guyon}, \binits{I.}},
\beditor{\bsnm{Luxburg}, \binits{U.V.}},
\beditor{\bsnm{Bengio}, \binits{S.}},
\beditor{\bsnm{Wallach}, \binits{H.}},
\beditor{\bsnm{Fergus}, \binits{R.}},
\beditor{\bsnm{Vishwanathan}, \binits{S.}},
\beditor{\bsnm{Garnett}, \binits{R.}} (eds.)
\bbtitle{Advances in Neural Information Processing Systems},
vol. \bseriesno{30}.
\bpublisher{Curran Associates, Inc.},
\blocation{Long Beach, CA, USA}
(\byear{2017}).
\burl{https://proceedings.neurips.cc/paper_files/paper/2017/file/8a20a8621978632d76c43dfd28b67767-Paper.pdf}
\end{bchapter}
\endbibitem

%%% 54
\bibitem[\protect\citeauthoryear{Ribeiro et~al.}{2016}]{ribeiro2016why}
\begin{bchapter}
\bauthor{\bsnm{Ribeiro}, \binits{M.T.}},
\bauthor{\bsnm{Singh}, \binits{S.}},
\bauthor{\bsnm{Guestrin}, \binits{C.}}:
\bctitle{Why should i trust you?: Explaining the predictions of any
  classifier}.
In: \bbtitle{Proceedings of the 22nd ACM SIGKDD International Conference on
  Knowledge Discovery and Data Mining}.
\bsertitle{KDD '16},
pp. \bfpage{1135}--\blpage{1144}.
\bpublisher{Association for Computing Machinery},
\blocation{New York, NY, USA}
(\byear{2016}).
\doiurl{10.1145/2939672.2939778} .
\burl{https://doi.org/10.1145/2939672.2939778}
\end{bchapter}
\endbibitem

%%% 55
\bibitem[\protect\citeauthoryear{Haeupler
  et~al.}{2014}]{haeupler2014consistent}
\begin{botherref}
\oauthor{\bsnm{Haeupler}, \binits{B.}},
\oauthor{\bsnm{Manasse}, \binits{M.}},
\oauthor{\bsnm{Talwar}, \binits{K.}}:
Consistent Weighted Sampling Made Fast, Small, and Easy
(2014).
\url{https://arxiv.org/abs/1410.4266}
\end{botherref}
\endbibitem

%%% 56
\bibitem[\protect\citeauthoryear{Guillot~Suarez}{2022}]{guillot2022human}
\begin{botherref}
\oauthor{\bsnm{Guillot~Suarez}, \binits{C.}}:
Human-in-the-loop hyperparameter tuning of deep nets to improve explainability
  of classifications.
Master's thesis,
Aalto University. School of Electrical Engineering
(2022).
\url{http://urn.fi/URN:NBN:fi:aalto-202205223354}
\end{botherref}
\endbibitem

%%% 57
\bibitem[\protect\citeauthoryear{Mosqueira-Rey
  et~al.}{2022}]{mosqueira2022classification}
\begin{bchapter}
\bauthor{\bsnm{Mosqueira-Rey}, \binits{E.}},
\bauthor{\bsnm{Pereira}, \binits{E.H.}},
\bauthor{\bsnm{Alonso-R\'{\i}os}, \binits{D.}},
\bauthor{\bsnm{Bobes-Bascar\'{a}n}, \binits{J.}}:
\bctitle{A classification and review of tools for developing and interacting
  with machine learning systems}.
In: \bbtitle{Proceedings of the 37th ACM/SIGAPP Symposium on Applied
  Computing}.
\bsertitle{SAC '22},
pp. \bfpage{1092}--\blpage{1101}.
\bpublisher{Association for Computing Machinery},
\blocation{New York, NY, USA}
(\byear{2022}).
\doiurl{10.1145/3477314.3507310} .
\burl{https://doi.org/10.1145/3477314.3507310}
\end{bchapter}
\endbibitem

%%% 58
\bibitem[\protect\citeauthoryear{Anderson et~al.}{2016}]{anderson2016runtime}
\begin{barticle}
\bauthor{\bsnm{Anderson}, \binits{M.R.}},
\bauthor{\bsnm{Antenucci}, \binits{D.}},
\bauthor{\bsnm{Cafarella}, \binits{M.J.}}:
\batitle{Runtime support for human-in-the-loop feature engineering system.}
\bjtitle{IEEE Data Eng. Bull.}
\bvolume{39}(\bissue{4}),
\bfpage{62}--\blpage{84}
(\byear{2016})
\end{barticle}
\endbibitem

%%% 59
\bibitem[\protect\citeauthoryear{Gkorou et~al.}{2020}]{gkorou2020get}
\begin{bchapter}
\bauthor{\bsnm{Gkorou}, \binits{D.}},
\bauthor{\bsnm{Larranaga}, \binits{M.}},
\bauthor{\bsnm{Ypma}, \binits{A.}},
\bauthor{\bsnm{Hasibi}, \binits{F.}},
\bauthor{\bsnm{Wijk}, \binits{R.J.}}:
\bctitle{Get a human-in-the-loop: Feature engineering via interactive
  visualizations}.
In: \bbtitle{Proceedings of the Workshop on Interactive Adaptive Learning
  Co-located with European Conference on Machine Learning and Principles and
  Practice of Knowledge Discovery in Databases (ECML PKDD 2020)},
vol. \bseriesno{2660}.
\bpublisher{CEUR Workshop Proceedings},
\blocation{Ghent, Belgium}
(\byear{2020}).
\burl{https://ceur-ws.org/Vol-2660/ialatecml_shortpaper4.pdf}
\end{bchapter}
\endbibitem

%%% 60
\bibitem[\protect\citeauthoryear{Rudin}{2019}]{rudin2019stop}
\begin{barticle}
\bauthor{\bsnm{Rudin}, \binits{C.}}:
\batitle{Stop explaining black box machine learning models for high stakes
  decisions and use interpretable models instead}.
\bjtitle{Nature machine intelligence}
\bvolume{1}(\bissue{5}),
\bfpage{206}--\blpage{215}
(\byear{2019})
\doiurl{10.1038/s42256-019-0048-x}
\end{barticle}
\endbibitem

\end{thebibliography}

\end{document}